%% file: main.tex
\documentclass[letterpaper]{article} 
\usepackage[preprint]{aaai2027}  
\usepackage[hyphens]{url}  
\usepackage{graphicx} 
\usepackage{natbib}  
\usepackage{caption} 
\usepackage{subcaption} 
\usepackage{algorithm}
\usepackage{algpseudocode}
\usepackage{enumitem}

\usepackage{newfloat}
\usepackage{listings}
\DeclareCaptionStyle{ruled}{labelfont=normalfont,labelsep=colon,strut=off} 
\floatstyle{ruled}
\newfloat{listing}{tb}{lst}{}
\floatname{listing}{Listing}

\usepackage{booktabs}

\usepackage{xspace}

\usepackage{amsmath}
\usepackage{mathtools} 
\usepackage{amssymb}
\usepackage{amsthm}
\usepackage{xspace}
\usepackage{cleveref}   
\usepackage{tikz}
\usetikzlibrary{positioning,calc,fit,backgrounds,arrows.meta,shapes.callouts,patterns}
\usepackage{pgfplots}
\usepackage{pgfplotstable} 
\pgfplotsset{compat=1.17}

\input{figures/leanstyle.tex}

\input{macros.tex}   

\title{Does the Proof Prove It That Way? Faithful Formalization of Elements Proofs}

\author{
    Tadd Mao\textsuperscript{\rm 1},
    Tianjun Zhong\textsuperscript{\rm 2},
    Dhruva Arekar\textsuperscript{\rm 2},
    Yuming Feng\textsuperscript{\rm 2},
    One An\textsuperscript{\rm 3},\\
    Jiani Huang\textsuperscript{\rm 3},
    Xujie Si\textsuperscript{\rm 1}\equalcontrib,
    Ziyang Li\textsuperscript{\rm 2}\equalcontrib\corresponding
}
\affiliations{
    \textsuperscript{\rm 1}University of Toronto\\
    \textsuperscript{\rm 2}Johns Hopkins University\\
    \textsuperscript{\rm 3}University of Pennsylvania\\
    ziyang@cs.jhu.edu
}

\begin{document}
\maketitle

\input{sections/1-abs.tex}

\input{sections/2-intro.tex}

\input{sections/3-motivating.tex}

\input{sections/4-methodology.tex}

\input{sections/5-evaluation.tex}

\input{sections/6-related.tex}

\input{sections/7-conclusion.tex}

\section*{Acknowledgements}
We thank the anonymous reviewers who participated in our human evaluation study.
Research reported in this publication was supported by an Amazon Research Award, Fall 2025, and by the 2026 Amazon Nova AI Challenge: Trusted Software Agents;
Tadd Mao acknowledges support from an NSERC Undergraduate Student Research Award.
The views and conclusions contained herein are those of the authors and should not be interpreted as necessarily representing the official policies of the supporting sponsors.

\bibliography{aaai2027}

\clearpage

\input{sections/8-appendix.tex}

\end{document}

%% file: figures/leanstyle.tex
\definecolor{leanKw}{RGB}{37,66,128}     
\definecolor{leanTac}{RGB}{17,122,101}   
\definecolor{leanStr}{RGB}{150,95,20}    
\definecolor{leanCmt}{RGB}{130,130,130}  
\lstdefinelanguage{Lean}{%
  morekeywords=[1]{theorem,lemma,def,by,have,show,from,fun,match,with,intro,%
    intros,exact,apply,obtain,wlog,generalizing,namespace,end,import,%
    set_option,as,at},
  morekeywords=[2]{euclid_intros,euclid_intro_sentence,euclid_sentence,%
    euclid_segment,euclid_conclude_sentence,euclid_apply,euclid_assumption,%
    by_contra,by_cases,assumption},
  morekeywords=[3]{euclid_finish},
  morekeywords=[4]{Point,Line,Prop,Bool,True,False,Triangle},
  alsoletter={_},
  sensitive=true,
  morestring=[b]{"},
  morecomment=[l]{--},
  morecomment=[s]{/-}{-/},
}
\lstdefinestyle{lean}{%
  language=Lean,
  basicstyle=\ttfamily\fontsize{7.5}{7.8}\selectfont,
  keywordstyle=[1]\color{leanKw}\bfseries,
  keywordstyle=[2]\color{leanTac},
  keywordstyle=[3]\color{red!75!black}\bfseries,
  keywordstyle=[4]\color{leanKw!70},
  stringstyle=\color{leanStr},
  commentstyle=\color{leanCmt}\itshape,
  numbers=none, xleftmargin=0pt, xrightmargin=0pt,
  aboveskip=1pt, belowskip=1pt,
  showstringspaces=false, breaklines=true, breakatwhitespace=true,
  columns=fullflexible, keepspaces=true,
  extendedchars=true, inputencoding=utf8,
  literate=%
    {∀}{{$\forall$}}1 {∃}{{$\exists$}}1 {∠}{{$\angle$}}1 {∟}{{$\llcorner$}}1 {─}{{\textendash}}1
    {≠}{{$\neq$}}1 {→}{{$\rightarrow$}}1 {∧}{{$\wedge$}}1 {∨}{{$\vee$}}1
    {¬}{{$\neg$}}1 {△}{{$\triangle$}}1 {≤}{{$\leq$}}1 {≥}{{$\geq$}}1
    {∈}{{$\in$}}1 {⟨}{{$\langle$}}1 {⟩}{{$\rangle$}}1 {·}{{$\cdot$}}1
    {…}{{$\dots$}}1 {⋮}{{$\vdots$}}1
    {α}{{$\alpha$}}1 {β}{{$\beta$}}1 {φ}{{$\phi$}}1 {κ}{{$\kappa$}}1 {Δ}{{$\Delta$}}1
    {ℝ}{{$\mathbb{R}$}}1 {∪}{{$\cup$}}1 {⊢}{{$\vdash$}}1 {∅}{{$\varnothing$}}1 {←}{{$\leftarrow$}}1
    {′}{{$'$}}1 {₁}{{$_1$}}1 {₂}{{$_2$}}1 {₃}{{$_3$}}1 {₄}{{$_4$}}1 {ₜ}{{$_t$}}1 {ᵢ}{{$_i$}}1
    {ₖ}{{$_k$}}1 {ₙ}{{$_n$}}1 {₋}{{$_-$}}1
}

%% file: macros.tex
\newcommand{\methodname}{\textsc{Pistis}\xspace}   
\newcommand{\fillalgoname}{\textsc{\textsc{OrderDecompose}}\xspace}   

\newcommand{\mapstage}{map stage\xspace}
\newcommand{\fillstage}{fill stage\xspace}
\newcommand{\script}{script\xspace}
\newcommand{\llm}{\ensuremath{\mathcal{M}}\xspace}
\newcommand{\oracle}{oracle\xspace}

\newcommand{\faithfulcriteria}{five necessary conditions of faithfulness\xspace}
\newcommand{\NL}{\ifmmode \text{NL}\else NL\xspace\fi}
\newcommand{\Formal}{\ifmmode \text{Lean}\else Lean\xspace\fi}

\newcommand{\coverage}{\textsc{Coverage}\xspace}
\newcommand{\atomicity}{\textsc{Atomicity}\xspace}
\newcommand{\atomicfaithfulness}{\textsc{Atomic Faithfulness}\xspace}
\newcommand{\order}{\textsc{Order}\xspace}
\newcommand{\citcond}{\textsc{Citation}\xspace}  
\newcommand{\soundness}{\textsc{Soundness}\xspace}

\newcommand{\assert}[1]{\operatorname{assert}(#1)}            
\newcommand{\assumptions}[1]{\operatorname{assumps}(#1)}  

\newcommand{\propN}{\ensuremath{P_\texttt{NL}}}
\newcommand{\PropNProof}{\ensuremath{F_\texttt{NL}}}

\newcommand{\numAssumpTags}{311\xspace}
\newcommand{\numAssumpGapTags}{33\xspace}
\newcommand{\numTotalEuclidGaps}{17\xspace} 
\newcommand{\numSuppressCiteTags}{7\xspace} 

\newcommand{\numAssumpGapTagsBookI}{11\xspace}
\newcommand{\numAssumpGapTagsBookII}{5\xspace}
\newcommand{\numAssumpGapTagsBookIII}{17\xspace}

\newcommand{\timesFasterThanLeanEuclid}{33}

\newcommand{\formal}[1]{\operatorname{formal}(#1)}

\newcommand{\numproofs}{92\xspace}
\newcommand{\closes}[2]{\Delta \cup \{#1\} \vdash_{\text{SMT-30s}} #2}

\newcommand{\sufficient}{\operatorname{SF}}
\newcommand{\suppliable}{\operatorname{SP}}

\newcommand{\hypothesis}{\alpha}
\newcommand{\conclusion}{\beta}
\newcommand{\systemEAxioms}{\Delta}

\crefname{figure}{Fig.}{Figs.}
\Crefname{figure}{Fig.}{Figs.}
\crefname{table}{Tab.}{Tabs.}
\Crefname{table}{Tab.}{Tabs.}
\crefname{section}{Sec.}{Secs.}
\Crefname{section}{Sec.}{Secs.}
\crefname{algorithm}{Alg.}{Algs.}
\Crefname{algorithm}{Alg.}{Algs.}

\definecolor{cBaseline}{RGB}{46,110,178}   
\definecolor{cBaselineD}{RGB}{27,71,122}   
\definecolor{cMine}{RGB}{240,160,40}       
\definecolor{cMineD}{RGB}{176,102,10}      
\pgfplotsset{
  barbase/.style={draw=cBaselineD, line width=0.5pt, fill=cBaseline, fill opacity=0.35},
  barmine/.style={draw=cMineD,     line width=0.5pt, fill=cMine,     fill opacity=0.45},
}

\definecolor{sA}{RGB}{30,90,180}    
\definecolor{sB}{RGB}{30,130,60}    
\definecolor{sC}{RGB}{200,110,20}   
\definecolor{sD}{RGB}{150,50,150}   
\newsavebox{\cnbox}
\newcommand{\cnum}[2]{%
  \sbox{\cnbox}{\bfseries\scriptsize #2}%
  \tikz[baseline=(cn.base)]{%
    \node[circle, fill=#1, text=white, font=\bfseries\scriptsize,
          inner sep=0.3pt, minimum size=7.5pt] (cn)
      {\ifdim\wd\cnbox>5pt\relax\scalebox{0.68}{#2}\else#2\fi};}}
\newcommand{\smark}[2][black!60]{\,\cnum{#1}{#2}}

\newcommand{\refpill}[1]{\tikz[baseline=(rp.base)]{%
  \node[inner xsep=3pt, inner ysep=0.6pt, rounded corners=2pt,
        fill=blue!8, draw=blue!40, text=blue!50!black,
        font=\scriptsize] (rp) {#1};}}

%% file: sections/1-abs.tex
\begin{abstract}

In formal verification, both the autoformalization of statements and automated proof search have been studied extensively.
While automated proof search can produce a formal proof that compiles, the generated proof does not necessarily reflect how the natural-language argument arrives at its conclusion---a property we refer to as \emph{faithfulness}.
With faithfully formalized proofs, one can check the reasoning behind a human- or AI-written argument, and assist mathematicians in formalizing their proof sketches.
However, it is particularly challenging due to misalignment of formal proof tactics and natural language reasoning.
In this work, we rigorously describe a set of five necessary conditions a faithful formal proof must satisfy, and introduce \textsc{Pistis}, an agentic, oracle-guided proof search that produces formal Lean proofs that satisfy them.
At its core is a novel faithfulness-preserving divide-and-conquer search, which we name \textsc{OrderDecompose}, that tracks citation dependencies and blocks unfaithful shortcuts, paired with a refutation search, that surfaces gaps and errors in the natural language proof source.
\textsc{OrderDecompose} completes proofs that baselines cannot close even within a 12-hour budget, and its artifacts compile over \timesFasterThanLeanEuclid$\times$ as fast as prior work's.
We apply \textsc{Pistis} on the first three books of Euclid's \emph{Elements}, producing high-quality artifacts containing faithful formal proofs.
Under a blinded human study and an LLM-as-a-judge protocol on rigorous rubrics, \textsc{Pistis}-generated proofs are favored over prior works---2.89$\times$ and 5.2$\times$ as often by human reviewers and the LLM judge, respectively.
It further uncovers gaps in Euclid's proofs and their translation, and can accept or refute natural language proofs written by humans or AI, demonstrating that faithful formalization is useful as a proof-checking tool.

\end{abstract}


%% file: sections/2-intro.tex
\section{Introduction}
\label{sec:introduction}

\begin{quote}
  \small
We are not trying to meet some abstract production quota of definitions, theorems and proofs. The measure of our success is whether what we do enables people to understand and think more clearly and effectively about mathematics.
\begin{flushright}On Proof and Progress in Mathematics~\cite{thurston1994proof}\end{flushright}
\end{quote}

\noindent \emph{Autoformalization} translates natural-language (\NL) mathematics into the formal language of a proof assistant such as Lean or Rocq, where it can be checked mechanically~\cite{wu2022autoformalization,demoura2021lean}.
Most work targets the formalization of \emph{statements}~\cite{li2024survey,azerbayev2023proofnet}, yet the formalization of \emph{proofs} is often left to automated proof search, which hunts for \emph{a} sequence of proof tactics that closes the goal.
As a result, proof \emph{faithfulness}, whether the formal proof reflects how the \NL argument reaches its conclusion~\cite{avigad2009formal,proofflow}, is largely overlooked.


\input{figures/fig-motivation-img}

Faithfulness is not a stylistic nicety; it is what makes a formal proof useful beyond a yes or no verdict.
A faithful formalization lets us \emph{check correctness at the level of the argument}, certifying that the reasoning as written in \NL is valid.
When the argument has a hole, a faithful attempt \emph{localizes} it rather than silently papering over it with proof automation, letting us \emph{validate or refute} a given \NL proof.
It helps \emph{mathematicians} turn informal sketches into formal artifacts that still resemble what they wrote, and it keeps those artifacts \emph{legible} for study and teaching, since each \NL reasoning step maps directly to a formal step.
For a proof to contribute to the broader field, legibility is only the starting point; the community must also recognize, digest, and build upon its argument.
A human author can convey the insights behind a proof, while current AI tools remain black boxes~\cite{tao2026ai}.

Achieving faithfulness is hard for three reasons.
First, the target is intrinsically misaligned with the source:
a proof assistant like Lean or Rocq advances by \emph{tactics}, which transform proof states, whereas an \NL proof advances by \emph{sentences} of mathematical reasoning.
Second, \NL proofs may omit steps a human finds obvious but the formal system explicitly demands, so a faithful proof must \emph{fill these gaps} without altering mathematical proof structures.
Third, the field lacks a precise, checkable notion of what proof faithfulness \emph{is}~\cite{murphy_autoformalizing_2024,proofflow}: prior evaluations lean on coarse proxies such as LLM-as-a-judge semantic matching~\cite{lu2024formalalign,poiroux2025reliable,yu2025mathesis}, which cannot serve as a certification.

In this work we introduce \methodname\footnote{From the Greek \emph{pistis} ($\pi\acute{\iota}\sigma\tau\iota\varsigma$), meaning ``faith'' or ``trust.''}, a framework that formalizes an \NL proof into a faithful Lean proof, which we demonstrate in Euclidean geometry.
Our key insight is that faithfulness becomes tractable once the proof is \emph{segmented}: rather than searching for a single proof term that closes the whole theorem, we split the \NL proof into structural formal sub-goals.
This process is accomplished by the \textit{Map Stage} of \methodname, where an agent works with an oracle to ensure the faithfulness of the mathematical proof structure.

After mapping a faithful proof structure, \methodname invokes a \emph{Fill Stage} algorithm, \fillalgoname, which is a novel proof search backed by LLM agents.
Its key component is an \emph{ordered, hierarchical decomposition} that closes stratified sub-goals individually rather than the theorem as a whole, giving both efficient compilation and a tight feedback loop.
This is further accompanied by
1) a proof-structure preservation mechanism that guards against the rewriting of the top-level faithful structure,
2) agentic tools that enforce faithful citation dependencies so each supporting proposition is applied exactly where the argument invokes it, and
3) when a sub-goal will not close, refutation techniques such as proving the negated sub-goal or searching for a counterexample.

We evaluate \methodname on the first three books of Euclid's \emph{Elements} (\numproofs\ propositions), producing new Lean artifacts.
Beyond extending the previously formalized single book of LeanEuclid~\citep{murphy_autoformalizing_2024}, we argue our artifact is of higher quality.
They compile at least \timesFasterThanLeanEuclid $\times$ as fast,
and on four criteria across 127 blinded reviews, human evaluators favor \methodname-generated proofs 2.89$\times$ as often as the already faithfulness-claiming LeanEuclid.
An LLM-as-a-judge protocol scoring a rigorous five-dimension rubric agrees, favoring \methodname over LeanEuclid by a 26:5 win--loss ratio on Book~I.
An ablation shows that \fillalgoname finishes within a fixed budget, produces 2.5$\times$ as many completed proofs as a state-of-the-art LLM-agent baseline while being 80\% more cost-effective.
Finally, \methodname faithfully surfaces gaps and refutes erroneous \NL proofs, including 2 genuine citation mistakes in the \emph{Elements} translated text.

\paragraph{Contributions.}
In summary, we contribute:
\begin{itemize}
  \item new Lean artifacts of faithfully formalized proofs for Books~I--III of Euclid's \emph{Elements};
  \item a taxonomy of \faithfulcriteria a formal proof must satisfy;
  \item \methodname, an oracle-guided methodology for faithful proof formalization, built on the \fillalgoname search;
  \item a thorough evaluation on faithfulness, cost efficiency, and compilation time, including a double-blind human expert review, showing improvements over prior methods; and
  \item the identification of at least 52 gaps in \emph{Elements} proofs including 2 genuine mistakes in the translation.
\end{itemize}



%% file: figures/fig-motivation-img.tex
\begin{figure*}[ht]
\centering
\includegraphics{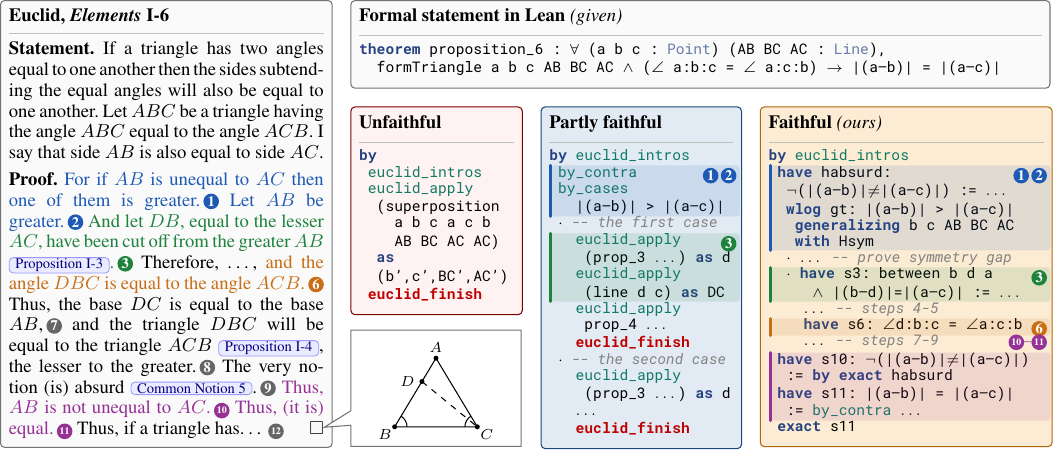}
\caption{
  Motivating example (Book~I, Prop.~6).
  \textbf{Left:} Euclid's statement and natural-language proof, its sentences marked \smark{1}--\smark{12} at the points where our method splits them;
  citations appear as pills (e.g.\ \refpill{Proposition I-3});
  the accompanied diagram illustrates the corresponding geometric constructions.
  \textbf{Top right:} the formal statement we are given in Lean.
  \textbf{Bottom right:} three ways to fill the proof, from unfaithful to faithful.
  Left: a single \texttt{superposition} that discards Euclid's argument;
  middle: prior work~\cite{murphy_autoformalizing_2024}, where one \texttt{euclid\_finish} collapses many of Euclid's sentences;
  and right: our \emph{faithful} proof, where each NL sentence maps to exactly one formal step matching \smark{k} on the left.
}
\label{fig:motivation}
\end{figure*}

%% file: sections/3-motivating.tex
\section{Motivating Example}
\label{sec:motivating}

Consider Proposition~6 of Book~I (I-6): if a triangle has two equal angles, the sides subtending them are equal (\cref{fig:motivation}).
Problem-wise, we are given its natural-language (\NL) proof as a string and its statement already formalized in Lean; our task is to fill in the formal proof.
It appears early in Euclid's \emph{Elements} and reads as a short, simple argument.
However, formalizing it \emph{faithfully} already forces us to discharge an implicit without-loss-of-generality step and several unstated deductions, difficulties that compound in later propositions.

\input{figures/conditions}

\paragraph{Faithfulness.}
Lean's kernel enforces \emph{proof irrelevance}: any two proofs of the same proposition are definitionally equal~\citep{carneiro2019typetheory}, so Lean accepts any term whose type is the stated proposition.
Therefore, all three proofs in \cref{fig:motivation} are treated as interchangeable.
Yet they differ sharply in how, and whether, they follow Euclid.
We call a formal proof \emph{faithful} when it mirrors the \NL proof step by step, each formal step corresponding to a specific \NL sentence, in the same order, under the assumptions that sentence actually invokes.

\paragraph{What a faithful proof for I-6 must capture.}
Euclid argues by contradiction.
A faithful formalization must therefore
(a)~follow this contradiction path, including the construction of $D$;
(b)~apply each cited proposition exactly where Euclid applies it, e.g.\ Proposition~I-3 for the construction and I-4 for the triangle congruence;
(c)~discharge the gaps Euclid leaves implicit, in particular the without-loss-of-generality (WLOG) step his ``one of them is greater.\smark[sA]{1} Let $AB$ be greater.\smark[sA]{2}'' leaves unstated; and
(d)~justify each step from precisely the claims that step is entitled to use.

\paragraph{Unfaithful and partly faithful proofs.}
The leftmost proof (\emph{unfaithful}) reflects triangle $ABC$ across the axis through $A$, producing a copy that is congruent, whence $AB=AC$.
This is a valid proof, but it proves by symmetry and reconstructs none of his steps.
The middle proof (\emph{partly faithful}), presented in prior work \citep{murphy_autoformalizing_2024}, recovers the high-level shape,
but three problems remain.
First, where Euclid reasons WLOG, it instead proves both cases directly.
Second, many sentences have no formal counterpart: sentence~\smark[sC]{6}, for instance, is absent, and one can only guess the correspondence by inspecting the referenced symbols.
Third, it may hide \NL reasoning steps in a black-box \texttt{euclid\_finish} tactic, which might take a different route than Euclid's intention and is non-deterministic.


\paragraph{A faithful proof.}
Our proof (\emph{faithful}, right) follows the \NL reasoning line by line.
Reading Euclid's ``the very notion is absurd''~\smark{9}, we recognize an argument by contradiction and name the negated goal \texttt{habsurd}.
Euclid never states WLOG; we infer it from ``one of them is greater.\smark[sA]{1} Let $AB$ be greater.\smark[sA]{2}'' and make it explicit with the \texttt{wlog} tactic, \emph{proving} the required symmetry rather than expanding both cases\footnote{Without-loss-of-generality (WLOG) is a well-known source of subtle errors, since the appealed-to symmetry is not always genuine.}.
This discharges the symmetry gap Euclid leaves implicit; had that symmetry been false, the attempt would instead expose the gap or refute the proof outright (\cref{sec:eval-accept-refute}).
We then proceed sentence by sentence, introducing each assertion as a named subgoal with \texttt{have} (\texttt{stepK}, shown \texttt{sK} in the figure), so sentences~\smark[sB]{3} and~\smark[sC]{6} map exactly.
Finally, mirroring Euclid's conclusion in sentences~\smark[sD]{10} and~\smark[sD]{11}, we discharge \texttt{habsurd} and close the proof with \texttt{exact}.

As we show later, the same faithful formalization, applied across the corpus, further lets us identify gaps in the \emph{Elements} and its translations (such as a reference to the wrong proposition), produce proofs that human readers prefer, and produce proofs that compile faster than automation-heavy alternatives.

%% file: figures/conditions.tex
\begin{table*}[t!]
\centering
\footnotesize
\caption{Conditions \methodname guarantees; all notation is defined in the Problem Definition. Faithfulness comprises the \faithfulcriteria; \soundness is separate. A free variable $s_i$ ranges over all sentences $s_1, \dots, s_n$. \script checks are automated (no \llm), \oracle checks are by human experts, and \soundness is checked by Lean.}
\label{tab:conditions}
\begin{tabular}{@{}l p{0.70\textwidth} l@{}}
\toprule
Name & Condition & Checked by \\
\midrule
\coverage           & $s_1,\dots,s_n$ concatenate, in order, to the NL proof $\propN$ & Script \\
\atomicity          & $s_i$ has a \emph{single} assertion $\assert{s_i}$, with assumptions $\assumptions{s_i}$ & Oracle \\
\atomicfaithfulness & $\forall \text{a} \in \assumptions{s_i}, \formal{\text{a}}$ is faithful, and $\formal{\assert{s_i}}$ is faithful  & Oracle \\
\order              & $(\formal{\assert{s_1}}, \dots, \formal{\assert{s_n}})$ is a subsequence of $(\phi_1, \dots, \phi_N)$ & Script \\
\citcond            & Given $\formal{\assert{s_i}}=\phi_{k}$, every (NL) prop. $p$ cited in $s_i$ has $\formal{p} \in \{\phi_j : j < k\}$ & Script \\
\midrule
\soundness          & $\formal{\propN}$ compiles in Lean with no \texttt{sorry} & Lean \\
\bottomrule
\end{tabular}
\end{table*}

%% file: sections/4-methodology.tex
\section{Methodology}

\label{sec:methodology}

In this section, we describe the problem definition of faithful formalization of \NL proofs, including one taxonomy of necessary faithful conditions (\cref{sec:methodology-problem}).
We then present the overview of \methodname, including its map stage, fill stage, and the main \textsc{OrderDecomposition} algorithm (\cref{sec:methodology-overview}).

\subsection{Problem Definition}
\label{sec:methodology-problem}

%
We begin by explaining what \methodname takes as input, what it produces as output, and what conditions these satisfy.



\paragraph{Input.}
\methodname takes as input $\langle \propN, \formal{\propN}, \PropNProof, \Delta \rangle$,
where $\propN$ is an \NL proposition, $\formal{\propN}$ is its formalization in Lean, and $\PropNProof$ is an \NL proof of $\propN$.
Specifically, $\formal{\propN}$ is of the form $\hypothesis \to \conclusion$, which reads ``given the hypotheses $\hypothesis$, the theorem concludes $\conclusion$.''
We may also write $\alpha=\alpha_1 \land ... \land \alpha_m$ to separate the larger formula of hypotheses into the individual hypothesis formulas $\alpha_1, \dots, \alpha_m$, and free variables in $\alpha$ are referred to as $v_1, ..., v_k$; the Appendix explains why these occur, and concrete examples in Lean.
Additionally, the set of System E \citep{avigad2009formal} axioms is denoted $\systemEAxioms$ and is also part of the input.

\paragraph{Output.}
While general proof search algorithms only output an arbitrary sound formal proof $\formal{\PropNProof}$, we argue that to ensure \emph{faithfulness}, the output requires more than just the formal proof itself.
In this work, we enforce a faithful proof to capture:
(i) knowing each step of the formal proof $\phi_1, ..., \phi_N$ (rather than the whole proof as one object $\phi$) so we can match these formal steps to the \NL steps,
(ii) what we even mean by an \NL step and how the NL proof is segmented into steps,
(iii) what is the structure within an \NL step, and finally
(iv) which NL step maps to which formal step.

Therefore, to address the above requirements, the output of \methodname is a tuple $\langle \bar{\phi}, \bar{s}, \assumptions{\cdot}, \assert{\cdot}, \formal{\cdot} \rangle$ where:
(1) $\bar{\phi} = \{\phi_i\}_1^N = \formal{\PropNProof}$ which is a sequence of formal steps,
with each $\phi_i$ derivable from the System E axioms $\systemEAxioms$, hypotheses $\hypothesis$ and $\phi_1, \dots, \phi_{i-1}$.
Formally $\bar{\phi}$ is a proof that $\systemEAxioms \cup \{\hypothesis\} \vdash \conclusion$;
(2) an NL partition of $\PropNProof$ into steps $\bar{s} = \{s_1, \dots, s_n\}$, where each \NL step $s_i$ is referred to as a \emph{segment};
(3) two mappings $\assumptions{\cdot}$ and $\assert{\cdot}$, sending each segment $s_i$ to a \emph{set} of substrings of $s_i$ in $\assumptions{s_i}$ and a single substring of $s_i$ in $\assert{s_i}$, respectively.
For example, consider $s_i = $ ``since $P$ and $Q$, therefore $R$'' we have $\assumptions{s_i} = \{P, Q\}$ and $\assert{s_i} = R$;
(4) a mapping $\formal{\cdot}$ that sends each assertion $\assert{s_i}$ to some $\phi_{j}$ and each assumption in $\assumptions{s_i}$ to some $\phi_{k}$.



\paragraph{Conditions.}
Whereas a typical proof search need only \emph{soundness}, i.e., Lean compilation, faithfulness demands more, and we present these additional conditions in \cref{tab:conditions}.
They are \emph{necessary} but not \emph{sufficient}: they constrain the proof's structure while leaving aspects such as author's implicit intent uncaptured.
Complete faithfulness is, we argue, unattainable: since a formal proof must fill gaps the \NL proof omits, a formalization mirroring the \NL proof exactly would leave those gaps and hence be unsound.
We therefore aim to be \emph{more} faithful than prior work rather than perfectly faithful, and leave this gap to future work.

\subsection{Approach Overview}
\label{sec:methodology-overview}

In order to produce a faithful formal proof satisfying the aforementioned conditions, \methodname uses a two-stage approach:
a \emph{map stage}, which creates a faithful and formal template with \texttt{sorry} placeholders,
and a \emph{fill stage}, which uses \fillalgoname to close the formal goal.

\input{figures/alg-orderdecompose}

\paragraph{Map Stage.}



$M:=\langle \bar{\phi'},\bar{s}, \assumptions{\cdot}, \assert{\cdot}, \formal{\cdot} \rangle$ are produced by the map stage, where $\bar{\phi'}$ is intended to be a subsequence of the (eventual) formal proof $\bar{\phi}$.
These outputs are created by LLM agents to ensure every condition in \cref{tab:conditions} except for \soundness, and \citcond, and are checked by an \oracle or a \script .
Details of the LLM-agent and its prompting strategies are described in the Appendix.

If the conditions the \oracle is assigned to check are not satisfied, the \oracle prompts the LLM to fix the output, and this process repeats until the \oracle is satisfied.

\paragraph{Fill stage.}





The fill stage proves each $\phi_i'$ since they were left unproven (\texttt{sorry}) in the map stage.
By ``proves`` we mean for each $\phi_i'$, create a sequence of steps $\kappa_1, ..., \kappa_t$ s.t. Lean compiler is able to check that $\systemEAxioms \cup \{\hypothesis, \phi'_{1}, ..., \phi'_{i-1}, \kappa_1, ..., \kappa_t\} \vdash \phi'_i$.
When the check succeeds, $\kappa_1, ..., \kappa_t$ are added to the proof: $(\phi'_1, ..., \phi'_{N'}) \gets (\phi'_1, ..., \phi'_{i-1}, \kappa_1, ..., \kappa_t, \phi'_i, ... \phi'_{N'})$.
This process is done using \fillalgoname (\cref{alg:orderdecompose}).

\fillalgoname takes in hypotheses $\hypothesis$, subgoals $\bar{\phi'}$, and conclusion $\conclusion$, along with an LLM $\mathcal{M}$.
It returns either \texttt{True} if it successfully proves all subgoals $\phi_i'$, or \texttt{False} if $\mathcal{M}$ decides to \textsc{GiveUp}.
The main structure of \fillalgoname is to prove $\phi'_1$, then $\phi'_2$, until $\phi'_{N'}$ in \emph{order} (\cref{alg:orderdecompose}, line 1--9), and for each $\phi_i'$, does a recursive, divide-and-conquer search by \emph{decomposing} it into further sub-lemmas (line 14--19) until each sub-lemma, is proven, resulting in $\phi_i'$ to be proven.
LLM executes and follows \fillalgoname in practice through skill files for general instructions, agentic hooks to prevent ``cheating'', and agentic tools to aid in the proof search (details in the Appendix).

The core of \fillalgoname is the recursive \textsc{Decomp} function (\cref{alg:orderdecompose}, line 10), which proves a subgoal $\nu$ from hypotheses $\eta_1, ..., \eta_n$ under a 30-second SMT budget, denoted $\closes{\cup_j \eta_j}{\nu}$; the tight budget keeps the feedback loop fast.
When a direct SMT search fails, \textsc{Decomp} asks the LLM to propose intermediate lemmas $h^{(j)} \to \omega^{(j)}$, keeps them only if they are \emph{sufficient} (SF) to close $\nu$ and \emph{suppliable} (SP) from the current context, and recurses on each.
In both \textsc{CreateLemmas} and \textsc{CreateHypothesis}, the lemma search is aided by agentic tools.
Finally, because the map stage leaves \soundness (P) and \citcond unverified, \fillalgoname checks both, and also that map stage objects were unchanged (line 7).

If \fillalgoname returns \texttt{False}, either:
(1) LLM is incapable
(2) The oracle made a map stage mistake
(3) System E implementation limitations
(4) Euclid made a mistake or
(5) The translated text made a mistake.
First, \fillalgoname's purpose is to mitigate LLM limitations.
Second, if the oracle makes a mistake, they fix it and \emph{resume} \fillalgoname (details in the Appendix),
and finally, (4) and (5) are formally addressed in the next section.




\paragraph{Acceptance, Refute, and Gaps.}
To address issues (4) and (5), \methodname can \emph{refute} a proof or identify its \emph{gaps}; we define these notions precisely.
To \emph{refute} a proof is to derive a logical contradiction from it.
A \emph{gap} is a missing step that does not itself create a contradiction, and whether an omission is large enough to count as a gap is inherently subjective.
To \emph{accept} a proof is to certify that it contains no contradiction, and is therefore valid, though it may still contain gaps.

\methodname identifies gaps in two ways.
First, during the fill stage the LLM may flag a suspected gap by returning \texttt{False}; if the \oracle agrees, we mark it (details on how we mark in the Appendix).
Second, \methodname detects \emph{assumption gaps} automatically, in an optional stage between mapping and filling.
A script materializes each formal assumption as a \texttt{have} statement and tries to close it with automated tactics, including SMT search; any unclosed assumption becomes a gap, marked with a \texttt{sorry} and added to the sequence $(\phi'_i)_i$ to be discharged.
We treat an unclosed assumption as a reliable gap because assumptions \emph{should already be provable}, so these tactics should close them.



\methodname \emph{refutes} a proof by formally exhibiting one of two failures at a step.
The first is that an asserted step is false under premises, $\exists i,\ \systemEAxioms \cup \{\hypothesis\} \vdash \neg \formal{\assert{s_i}}$: the axioms and hypotheses together \emph{disprove} what the proof claims at $s_i$.
The second is that an asserted step does not follow from the premises, $\exists i,\ \systemEAxioms \vdash \exists v_1, ..., v_k\, \neg (\hypothesis \to \formal{\assert{s_i}})$: there is a concrete configuration of the geometric objects $v_1, ..., v_k$ (the free variables of $\hypothesis$) that satisfies the hypotheses yet violates $\assert{s_i}$.
Either failure yields a contradiction that certifies the proof is wrong; we further detail how this is derived in the Appendix, and give examples in \cref{sec:eval-accept-refute}.


%% file: figures/alg-orderdecompose.tex
\begin{algorithm}[t!]
\small
\caption{\fillalgoname}
\label{alg:orderdecompose}
\definecolor{algcmt}{RGB}{20,110,60}
\renewcommand{\algorithmiccomment}[1]{\hfill{\color{algcmt}\itshape$\triangleright$~#1}}
\renewcommand{\algorithmicensure}{\textbf{Returns:}}
\begin{algorithmic}[1]
\Require Hypothesis $\tau$, Subgoals $\psi_1, \dots, \psi_n$ and conclusion $\gamma$ s.t. $\systemEAxioms \cup \{\tau, \psi_1, \dots, \psi_n\} \vdash \gamma$; LLM $\llm$
\Ensure \texttt{Bool} --- whether the proof search succeeded
\For{$i = 1, \dots, m$} \Comment{prove each subgoal $\psi_i$ in order}
  \Repeat \Comment{re-attempt $\psi_i$ until it passes every check}
  \State \textbf{if} $\textsc{GiveUp}(\tau, \psi_i, \llm)$ \textbf{then} \Return \texttt{False}
  \State $(\eta_1, ..., \eta_m) \gets \textsc{CreateHypothesis}(\tau, \psi_i, \llm)$
  \State $\suppliable \gets \bigwedge_j \textsc{InContext}(\eta_j)$ \Comment{suppliable}
  \State $\operatorname{P} \gets \textsc{Decomp}((\eta_1, ..., \eta_m), \psi_i, \llm)$ \Comment{soundness}
  \Until{$\suppliable \land \operatorname{P} \land \citcond(\psi_i) \land \textsc{UnChanged}(M \setminus \{\bar{\phi'}\})$}
\EndFor
\State \Return \texttt{True} \Comment{all subgoals proven, so $\gamma$ follows}
\Statex
\Function{Decomp}{$(\eta_1, ..., \eta_n), \nu, \llm$} : \texttt{Bool}
  \Repeat \Comment{close $\nu$ directly, else decompose it}
    \State \textbf{if} $\closes{\cup_j \{\eta_j\}}{\nu}$ \textbf{then} \Return \texttt{True}
    \State \textbf{if} $\textsc{GiveUp}(\bar{\eta}, \nu, \llm)$ \textbf{then} \Return \texttt{False}
    \Repeat \Comment{search lemmas until sufficient \& suppliable}
      \State $(h^{(j)} \to \omega^{(j)}) \gets \textsc{CreateLemmas}(\bar{\eta}, \nu, \llm)$
      \State $\sufficient \gets (\closes{\cup_k \eta_k \cup_{j} \omega^{(j)}}{\nu})$ \Comment{sufficient}
      \State $\forall j,(h^{(j)}_1, ..., h^{(j)}_{l_j}) \gets h^{(j)}$
      \State $\suppliable \gets \bigwedge_{j, i} \textsc{InContext}(h^{(j)}_i)$ \Comment{suppliable}
    \Until{$\sufficient \land \suppliable$}
    \State $\mathit{allHelpersClosed} \gets \bigwedge_j \textsc{Decomp}((h^{(j)}), \omega^{(j)}, \llm)$
  \Until{$\mathit{allHelpersClosed}$}
  \State \Return \texttt{True} \Comment{all helper lemmas proved, so $\nu$ holds}
\EndFunction
\end{algorithmic}
\end{algorithm}

%% file: sections/5-evaluation.tex
\input{figures/fig-survey-overview}

\section{Evaluation}
\label{sec:evaluation}

We conduct systematic evaluations of our method to answer the following four research questions:
\begin{enumerate}[leftmargin=0.115\linewidth,rightmargin=0\linewidth]
    \item[\textbf{RQ1.}] How faithful are the \methodname-generated formal proofs compared to LeanEuclid \citep{murphy_autoformalizing_2024}?
    \item[\textbf{RQ2.}] Is \fillalgoname necessary, or can a bare LLM complete the \emph{fill stage} on its own?
    \item[\textbf{RQ3.}] How does the compilation speed of \methodname-generated proofs compare to LeanEuclid?
    \item[\textbf{RQ4.}] Qualitatively, how does \methodname refute proofs and identify gaps, including those in the textbook translations?
\end{enumerate}

\subsection{Evaluation Setups}
\label{sec:eval-setups}

\paragraph{Targets.}
We target Euclid's \emph{Elements}, Books~I--III, comprising \numproofs propositions in total (48 in Book~I, 14 in Book~II, 30 in Book~III).
We take the natural-language proof text (including its bracketed proposition references) from open source \citet{fitzpatrick_rfitzpelements_2026}, and treat it as the source we formalize faithfully.
A small number of Book~III propositions are omitted as they lie beyond the scope of the current System E implementation.

\paragraph{Model and Agent.}
All proofs are generated with Claude Code (version 2.1.168), driving Claude Opus~4.8.

\paragraph{Machine configuration.}
Proofs are checked with Lean~4 (\texttt{v4.8.0-rc2}), using SMT solvers Z3~4.15.4 and cvc5~1.3.4. In the Lean compile-time experiments, the SMT solvers run uncapped, so the only cutoff is a wall-clock limit. Experiments were generally run on a Linux SLURM cluster whose compute nodes are Intel Xeon Platinum~8480+ (Sapphire Rapids), with each job allocated 32~cores and 128~GB.

\paragraph{Baselines and Metrics.}
We compare against LeanEuclid~\citep{murphy_autoformalizing_2024} (Book~I only), which claims to already provide faithful formalizations, and vanilla LLM-agent based baselines.
We report faithfulness (detailed in RQ1), wall-clock time, monetary cost, and other relevant metrics detailed in the following sections.

\input{figures/fig-survey-prop-collage-split}

\subsection{RQ1: Faithfulness of Generated Proofs}
\label{sec:eval-rq1-faithfulness}

\paragraph{Double-blind Human Evaluation.}
We randomly sampled 30 of the 48 Book~I propositions for human review.
Each reviewer was assigned 10 random propositions and saw the textbook theorem and proof, two anonymized formalizations, and an interactive Lean server for inspecting the proof state at each line.
The formalizations were shown in randomized order, and method-identifying nonmathematical artifacts were removed.
Reviewers were 14 formal-methods researchers and students familiar with Lean and geometry, who were independent of the author team.
The study produced 127 completed assignments, with 4.2 final annotations per proposition on average, covering 254 step-fidelity ratings and 381 pairwise preference judgments.
The review rubric and interface are detailed and summarized in the Appendix.

Figure~\ref{fig:survey-overview} summarizes the aggregate human scores and preferences, while Figure~\ref{fig:survey-prop-collage-split} shows the proposition-level pattern.
The human ratings favor our method across all evaluated dimensions.
Figure~\ref{fig:survey-overview} shows a higher mean step-fidelity score for our formalizations than for LeanEuclid (4.52 vs.\ 3.68).
Reviewers also preferred proofs generated by our method in 63.0\% of mathematical-transparency comparisons, 73.2\% of textbook-representation comparisons, and 62.2\% of overall-preference comparisons.
The proposition-level heatmaps in Figure~\ref{fig:survey-prop-collage-split} show that these gains are broadly distributed rather than driven by a small number of propositions.

\paragraph{LLM-as-Judge Evaluation.}
We also assess faithfulness with an LLM-as-judge protocol.
As shown in \cref{fig:llm-as-judge-radar}, our method consistently outperforms the baselines while staying consistent throughout the three books.
Specifically, faithfulness evaluation was done with a 5 point scale rubric across 5 different subcategories: object correspondence, proof step coverage, proof structure fidelity, cited dependency fidelity, and assumption and side conditions.
We prompted open and closed LLMs (Opus 4.6 shown, GLM-5 in Appendix) to perform assessment with a strict rubric, detailed in the Appendix.
On Book~I, \methodname is favored 5.2$\times$ as often as LeanEuclid (26 wins, 17 ties, 5 losses).


\input{data/llm_judge/scores.tex}

\input{figures/fig-radar} 

\subsection{RQ2: Filling Stage Ablation Study}
\label{sec:eval-rq2-ablation}
We compare \methodname (full pipeline) against an ablated \emph{bare-LLM} baseline that receives the same agentic framework and model but with our skills, scripts, hooks, and persistent memory removed.
Both arms start in the same state, where the map stage has just been completed.
Further, they are given the same citation instructions, are graded once at the end by an identical script checking \soundness, \citcond (\cref{tab:conditions}), and that none of the objects created in the map stage were altered.
They run under a 12-hour wall-clock budget per attempt, and we evaluate $15$ propositions across Books~I--III with $3$ repeated runs each (90 total runs).

\input{figures/fig-ablation-table} 

\input{figures/fig-ablation-wall} 

\paragraph{Findings.}
As shown in Table~\ref{tab:ablation}, and a more detailed view in \cref{fig:ablation}, \methodname solves all 15 propositions (45/45 runs succeed), whereas the baseline solves 6 and fails to finish the remaining 9 within the 12-hour budget (29/45 runs time out). \methodname is also cheaper per successful proof (\$15.6 vs.\ \$28.8), since the baseline pays for many long runs that never complete.

\subsection{RQ3: Lean Compile Performance}
\label{sec:eval-rq3-compile}
\input{figures/fig-compile-cactus}

Some LeanEuclid proofs compile \emph{very} slowly (over an hour) or not at all, due to heavy SMT reliance. A key design decision of \fillalgoname is to ensure proofs compile quickly, so here we compare \methodname generated proofs against the original LeanEuclid proofs~\citep{murphy_autoformalizing_2024} in terms of compilation speed on Book~I.
We wipe all built artifacts and compile\footnote{Compile-time
experiments were run on a single exclusive AMD~EPYC~9634 (Zen~4, 84~cores) in this specific experiment} every proposition from cold,
one at a time, with a one-hour per-proposition wall timeout, and repeat this over three runs.
As \cref{fig:compile-cactus} shows, \methodname compiles all $48$ propositions, whereas LeanEuclid compiles only $41$. Further, \methodname compiles
all of Book~I in $13.3$~minutes of wall-clock, more than $33\times$ as fast
as LeanEuclid ($7.3+$~hours).


\subsection{RQ4: Qual. Analysis of Refutations and Gaps}
\label{sec:eval-accept-refute}


We now illustrate how \methodname identifies gaps in Euclid's \emph{Elements} and its translations, or refutes \NL proofs such as those generated by AI.
Across the corpus of Books~I--III, we mark 33 assumption gaps, 2 confirmed citation mistakes, and 17 other generic gaps.
We hereby illustrate a citation gap found by \methodname arising from the edition rather than Euclid: \cref{fig:citation-mistake} shows a source that brackets ``let $AB$ be cut in half'' as [Prop.~I-9], though I-9 bisects an \emph{angle} whereas bisecting a straight-line is I-10.
Another example, an AI-generated proof refuted by our method, is presented in \cref{fig:accept-refute}, where the sentence $s_3$ of a proof of Book~I, Prop.~5 claims that angle $CBD$ is a right angle, which we formally refute.
Together, these demonstrate that \methodname is capable of checking \NL proofs while surfacing gaps and refuting errors.

\input{figures/fig-citation-mistake-img}
\input{figures/fig-accept-refute-img}

%% file: figures/fig-survey-overview.tex
\begin{figure}[t]
\centering
\pgfplotstableread{data/survey/overview_fidelity.dat}\surveyfid
\pgfplotstableread{data/survey/overview_preferences.dat}\surveypref
\begin{minipage}[b]{0.26\linewidth}
\centering
\begin{tikzpicture}
\begin{axis}[
    width=1\linewidth,
    scale only axis,
    height=2.1cm,
    ybar,
    bar width=9pt,
    ymin=0,
    ymax=5.4,
    title={Step Fidelity (0--5)},
    title style={font=\scriptsize, yshift=-1.2ex},
    xmin=-0.7,
    xmax=1.7,
    xtick={0,1},
    xtick pos=bottom,
    xticklabels={\strut LeanEuclid,\strut Ours},
    xticklabel style={font=\scriptsize},
    yticklabel style={font=\scriptsize, inner sep=0pt, xshift=-3pt},
    ytick={0,1,2,3,4,5},
    ymajorgrids,
    grid style={black!8},
    nodes near coords,
    every node near coord/.append style={font=\scriptsize, yshift=1pt, text=black, opacity=1},
]
\addplot[
    barbase,
    restrict x to domain=-0.1:0.1,
    bar shift=0pt,
    error bars/.cd,
    y dir=both,
    y explicit,
] table[x=x, y=mean, y error=se] {\surveyfid};
\addplot[
    barmine,
    restrict x to domain=0.9:1.1,
    bar shift=0pt,
    error bars/.cd,
    y dir=both,
    y explicit,
] table[x=x, y=mean, y error=se] {\surveyfid};
\end{axis}
\end{tikzpicture}
\end{minipage}
\hfill
\begin{minipage}[b]{0.71\linewidth}
\centering
\begin{tikzpicture}
\begin{axis}[
    width=0.66\linewidth,
    scale only axis,
    height=2.1cm,
    xbar stacked,
    xmin=0,
    xmax=100,
    bar width=7pt,
    xlabel={Preferred Responses (\%)},
    xlabel style={font=\scriptsize, yshift=3pt},
    ytick={0,1,2},
    yticklabels={{Mathematical\\[-3pt]transparency},{Textbook\\[-3pt]representation},{Overall\\[-3pt]preference}},
    yticklabel style={font=\scriptsize, align=right, text width=1.65cm},
    ytick style={draw=none},
    tick label style={font=\scriptsize},
    y dir=reverse,
    enlarge y limits=false,
    ymin=-0.75,
    ymax=2.55,
    legend style={
        font=\scriptsize,
        at={(0.5,1.02)},
        anchor=south,
        legend columns=3,
        draw=black!0,
        fill=white,
        /tikz/every even column/.append style={column sep=0.15em}
    },
    area legend,
    grid=major,
    grid style={black!8},
    every node near coord/.append style={font=\scriptsize, text=black, opacity=1, inner sep=1pt},
]
\addplot[barbase, nodes near coords={\pgfmathprintnumber[precision=0]{\pgfplotspointmeta}\%}] table[x=leanpct, y=y] {\surveypref};
\addplot[draw=black!45, line width=0.5pt, fill=black!14] table[x=tiepct, y=y] {\surveypref};
\addplot[barmine, nodes near coords={\pgfmathprintnumber[precision=0]{\pgfplotspointmeta}\%}] table[x=ourspct, y=y] {\surveypref};
\legend{LeanEuclid, Tie, Ours}
\end{axis}
\end{tikzpicture}
\end{minipage}
\caption{
    Human expert review results.
    Left: mean step-fidelity ratings with standard error over all rated items.
    Right: preference judgments by reviewers on three dimensions.
}
\label{fig:survey-overview}
\end{figure}
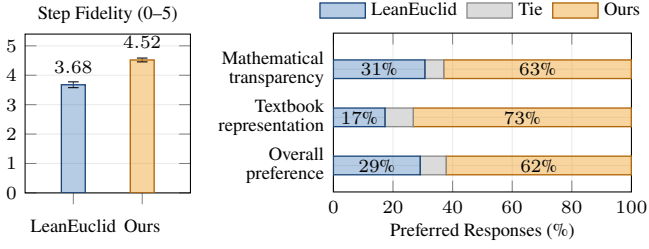

%% file: figures/fig-survey-prop-collage-split.tex
\begin{figure}[t]
\centering
\begin{tikzpicture}
\begin{axis}[
    name=leftsurvey,
    width=0.49\linewidth,
    height=0.20\textheight,
    xmin=-0.5,
    xmax=3.5,
    ymin=-0.5,
    ymax=14.5,
    y dir=reverse,
    axis line style={draw=none},
    tick style={draw=none},
    xtick={0,1,2,3},
    xticklabels={\strut Step, \strut Transp, \strut Repr, \strut Pref},
    xticklabel pos=upper,
    x tick label style={font=\scriptsize, align=center, yshift=-3pt},
    ytick={0,...,14},
    yticklabels={Prop I-1, Prop I-2, Prop I-6, Prop I-7, Prop I-8, Prop I-9, Prop I-11, Prop I-12, Prop I-13, Prop I-14, Prop I-15, Prop I-16, Prop I-17, Prop I-20, Prop I-23},
    y tick label style={font=\scriptsize},
    enlargelimits=false,
    grid=both,
    minor tick num=0,
    grid style={black!10},
    scatter,
    only marks,
    mark=square*,
    mark size=2.9pt,
    point meta min=-3,
    point meta max=3,
    colormap={surveyheat}{
        color(0cm)=(cBaseline);
        color(1cm)=(white);
        color(2cm)=(cMine)
    },
    scatter/use mapped color={draw=mapped color!65!black, fill=mapped color},
]
\addplot[scatter src=explicit] table[x=x, y=local_idx, meta=value] {data/survey/by_prop_split_cells_left.dat};
\end{axis}
\begin{axis}[
    name=rightsurvey,
    at={(leftsurvey.east)},
    anchor=west,
    xshift=0.06\linewidth,
    width=0.49\linewidth,
    height=0.20\textheight,
    xmin=-0.5,
    xmax=3.5,
    ymin=-0.5,
    ymax=14.5,
    y dir=reverse,
    axis line style={draw=none},
    tick style={draw=none},
    xtick={0,1,2,3},
    xticklabels={\strut Step, \strut Transp, \strut Repr, \strut Pref},
    xticklabel pos=upper,
    x tick label style={font=\scriptsize, align=center, yshift=-3pt},
    ytick={0,...,14},
    yticklabels={Prop I-25, Prop I-27, Prop I-28, Prop I-30, Prop I-32, Prop I-33, Prop I-35, Prop I-36, Prop I-37, Prop I-38, Prop I-39, Prop I-40, Prop I-42, Prop I-43, Prop I-46},
    yticklabel pos=right,
    y tick label style={font=\scriptsize},
    enlargelimits=false,
    grid=both,
    minor tick num=0,
    grid style={black!10},
    scatter,
    only marks,
    mark=square*,
    mark size=2.9pt,
    point meta min=-3,
    point meta max=3,
    colormap={surveyheat}{
        color(0cm)=(cBaseline);
        color(1cm)=(white);
        color(2cm)=(cMine)
    },
    scatter/use mapped color={draw=mapped color!65!black, fill=mapped color},
]
\addplot[scatter src=explicit] table[x=x, y=local_idx, meta=value] {data/survey/by_prop_split_cells_right.dat};
\end{axis}
\coordinate (splitbar) at ($(leftsurvey.south west)!0.5!(rightsurvey.south east)+(0,-0.22cm)$);
\shade[left color=cBaseline, right color=white]
    ($(splitbar)+(-0.38\linewidth,0)$) rectangle
    ($(splitbar)+(0,0.08cm)$);
\shade[left color=white, right color=cMine]
    ($(splitbar)+(0,0)$) rectangle
    ($(splitbar)+(0.38\linewidth,0.08cm)$);
\draw[black!45, line width=0.2pt]
    ($(splitbar)+(-0.38\linewidth,0)$) rectangle
    ($(splitbar)+(0.38\linewidth,0.08cm)$);
\node[font=\scriptsize, anchor=north] at ($(splitbar)+(-0.38\linewidth,-0.07cm)$) {LeanEuclid};
\node[font=\scriptsize, anchor=north] at ($(splitbar)+(0,-0.07cm)$) {Tie};
\node[font=\scriptsize, anchor=north] at ($(splitbar)+(0.38\linewidth,-0.07cm)$) {Ours};
\end{tikzpicture}
\caption{Per-proposition human evaluation on Book~I, each cell scored on the
$[-3,3]$ scale: blue favors LeanEuclid, white a tie, and gold favors ours.}
\label{fig:survey-prop-collage-split}
\end{figure}
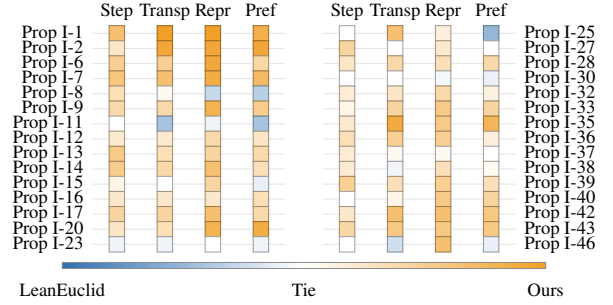

%% file: data/llm_judge/scores.tex

\def\radarDataNewOne{4.88,4.98,4.79,4.88,4.83}
\def\radarDataOldOne{4.71,4.56,4.50,4.79,4.71}
\def\radarDataNewTwo{5.00,5.00,4.86,4.93,4.86}
\def\radarDataOldTwo{3.86,1.14,0.79,0.57,4.50}
\def\radarDataNewThree{4.87,4.63,4.33,4.27,4.60}
\def\radarNOne{48}
\def\radarNTwo{14}
\def\radarNThree{30}

%% file: figures/fig-radar.tex
\colorlet{cRadarOurs}{cMine}      \colorlet{cRadarOursD}{cMineD}
\colorlet{cRadarBase}{cBaseline}  \colorlet{cRadarBaseD}{cBaselineD}
\definecolor{cRadarLLM}{HTML}{8E44AD}
\colorlet{cRadarLLMD}{cRadarLLM!70!black}
\newcommand{\radarpanel}[6]{%
  \edef\newdata{#1}%
  \foreach \val [count=\i from 0] in \newdata {%
    \expandafter\xdef\csname newval\i\endcsname{\val}%
  }%
  \def\hasold{0}%
  \def\testold{#2}%
  \ifx\testold\empty\else
    \def\hasold{1}%
    \edef\olddata{#2}%
    \foreach \val [count=\i from 0] in \olddata {%
      \expandafter\xdef\csname oldval\i\endcsname{\val}%
    }%
  \fi
  \resizebox{\linewidth}{!}{%
  \begin{tikzpicture}
    \def\R{1cm}
    \def\labels{{"OC","PSC","PSF","CDF","ASC"}}%
    \pgfmathsetmacro{\yscale}{1/(#4-#3)}
    \pgfmathtruncatemacro{\nticks}{#4-#3}%
    \foreach \lev in {1,...,\nticks} {%
      \pgfmathsetmacro{\r}{\lev/\nticks}%
      \draw[gray!30]
        (90:{\r cm}) \foreach \s in {1,...,4} {-- ({90-\s*72}:{\r cm})} -- cycle;
    }%
    \foreach \s in {0,...,4} {%
      \draw[gray!50] (0,0) -- ({90-\s*72}:\R);
    }%
    \node[font=\scriptsize] at (90:{\R+0.24cm}) {OC};
    \node[font=\scriptsize] at (42:{\R+0.14cm}) {PSC};
    \node[font=\scriptsize] at (-54:{\R+0.28cm}) {PSF};
    \node[font=\scriptsize] at (-126:{\R+0.28cm}) {CDF};
    \node[font=\scriptsize] at (138:{\R+0.14cm}) {ASC};
    \ifnum\hasold=1
      \draw[#6,thick,dashed,fill=#5,fill opacity=0.22]
        ({90}:{(\csname oldval0\endcsname-#3)*\yscale cm})
        \foreach \s in {1,...,4} {%
          -- ({90-\s*72}:{(\csname oldval\s\endcsname-#3)*\yscale cm})
        } -- cycle;
      \foreach \s in {0,...,4} {%
        \fill[#6] ({90-\s*72}:{(\csname oldval\s\endcsname-#3)*\yscale cm}) circle(1.3pt);
      }%
    \fi
    \draw[cRadarOursD,thick,fill=cRadarOurs,fill opacity=0.32]
      ({90}:{(\csname newval0\endcsname-#3)*\yscale cm})
      \foreach \s in {1,...,4} {%
        -- ({90-\s*72}:{(\csname newval\s\endcsname-#3)*\yscale cm})
      } -- cycle;
    \foreach \s in {0,...,4} {%
      \fill[cRadarOursD] ({90-\s*72}:{(\csname newval\s\endcsname-#3)*\yscale cm}) circle(1.3pt);
    }%
  \end{tikzpicture}%
  }%
}
\begin{figure}[t]
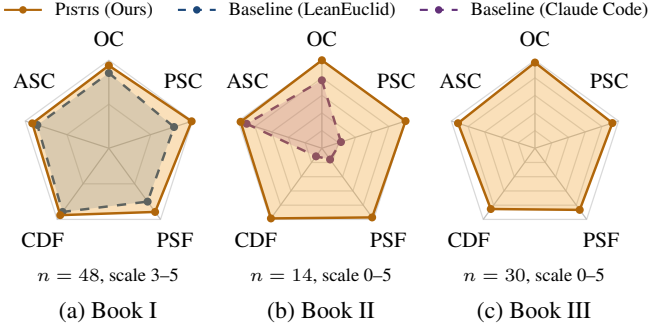

\centering
\tikz[baseline=-0.5ex]{%
  \draw[cRadarOursD,thick] (0,0) -- (0.5,0);
  \fill[cRadarOursD] (0.25,0) circle(1.4pt);
  \node[right,font=\scriptsize] at (0.55,0) {\methodname (Ours)};
  \draw[cRadarBaseD,thick,dashed] (2.3,0) -- (2.8,0);
  \fill[cRadarBaseD] (2.55,0) circle(1.4pt);
  \node[right,font=\scriptsize] at (2.85,0) {Baseline (LeanEuclid)};
  \draw[cRadarLLMD,thick,dashed] (5.5,0) -- (6.0,0);
  \fill[cRadarLLMD] (5.75,0) circle(1.4pt);
  \node[right,font=\scriptsize] at (6.05,0) {Baseline (Claude Code)};
}\\[-3pt]
\begin{subfigure}[b]{0.33\linewidth}
\centering
\radarpanel{\radarDataNewOne}{\radarDataOldOne}{3}{5}{cRadarBase}{cRadarBaseD}\\[-2pt]
{\scriptsize $n=\radarNOne$, scale 3--5}
\caption{Book~I}
\label{fig:radar-b1}
\end{subfigure}%
\hfill
\begin{subfigure}[b]{0.33\linewidth}
\centering
\radarpanel{\radarDataNewTwo}{\radarDataOldTwo}{0}{5}{cRadarLLM}{cRadarLLMD}\\[-2pt]
{\scriptsize $n=\radarNTwo$, scale 0--5}
\caption{Book~II}
\label{fig:radar-b2}
\end{subfigure}%
\hfill
\begin{subfigure}[b]{0.33\linewidth}
\centering
\radarpanel{\radarDataNewThree}{}{0}{5}{cRadarBase}{cRadarBaseD}\\[-2pt]
{\scriptsize $n=\radarNThree$, scale 0--5}
\caption{Book~III}
\label{fig:radar-b3}
\end{subfigure}
\caption{
  LLM-as-judge faithfulness by book over five subcategories,
  \textit{OC} (object correspondence), \textit{PSC} (proof step coverage), \textit{PSF} (proof structure fidelity), \textit{CDF} (cited dependency fidelity), \textit{ASC} (assumption \& side condition).
  Book~I compares \methodname against the LeanEuclid baseline, while Book~II compares against LLM-agent-generated proofs.
}
\label{fig:llm-as-judge-radar}
\end{figure}

%% file: figures/fig-ablation-table.tex
\begin{table}[t]
\centering
\footnotesize
\setlength{\tabcolsep}{2pt}
\begin{tabular}{@{}lccc@{}}
\toprule
\textbf{Metric} & \textbf{\fillalgoname} & \textbf{LLM-only} & \textbf{Factor} \\
\midrule
coverage ($\mathrm{pass@}k$) $\uparrow$   & \textbf{15/15 (100\%)} & 6/15 (40\%)  & $2.5\times$ \\
total cost / success (\$) $\downarrow$          & \textbf{15.6}  & 28.8  & $1.8\times$ \\
\bottomrule
\end{tabular}
\caption{Fill Stage ablation ($N{=}15$, $k{=}3$ runs each).
}
\label{tab:ablation}
\end{table}

%% file: figures/fig-ablation-wall.tex
%
\def\logsh{1.4}
\def\wallfrac{0.56}
\def\compfrac{0.44}
\def\timeouthrs{12} 
\pgfplotstableread{data/ablation/wall_by_prop.dat}\wall
\pgfplotstableread{data/ablation/compile/compile_by_prop.dat}\comp
\pgfplotsset{
  rq2axis/.style={
    height=0.37\textheight, xbar, bar width=1.7pt, bar shift auto=false,
    enlarge y limits={abs=0.8},
    ytick={0,1,2,3,4,5,6,7,8,9,10,11,12,13,14},
    y dir=reverse, ytick style={draw=none},
    tick label style={font=\scriptsize},
    xlabel style={font=\footnotesize},
    grid=major, grid style={black!8}, area legend,
  },
}
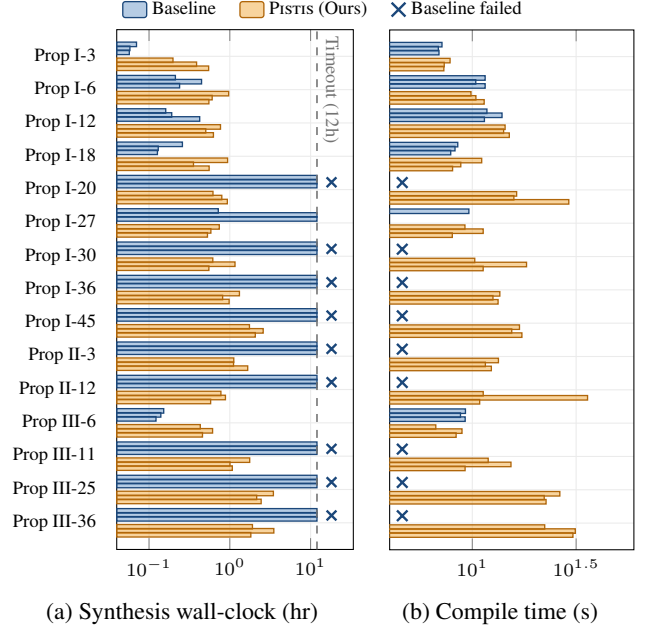
\begin{figure}[t]
\centering
\newcommand{\lgswatch}[3]{\tikz[baseline=-0.5ex]\node[draw=#1, line width=0.5pt,
    fill=#2, fill opacity=#3, minimum width=9pt, minimum height=6pt,
    inner sep=0pt]{};}
\newcommand{\lgx}{\tikz[baseline=-0.5ex]\draw[cBaselineD, line width=0.9pt]
    (-2.6pt,-2.6pt)--(2.6pt,2.6pt) (-2.6pt,2.6pt)--(2.6pt,-2.6pt);}
\tikz\node[fill=white, rounded corners=1pt, inner xsep=6pt,
    inner ysep=3pt, font=\scriptsize]{%
    \lgswatch{cBaselineD}{cBaseline}{0.35}~Baseline \quad
    \lgswatch{cMineD}{cMine}{0.45}~\methodname (Ours) \quad
    \lgx~Baseline failed};\\[0pt]
\begin{subfigure}[t]{\wallfrac\linewidth}
\centering
\begin{tikzpicture}
\pgfmathsetmacro{\toX}{ln(\timeouthrs)/ln(10)+\logsh}
\begin{axis}[rq2axis,
    width=\linewidth,
    xmin=0, xmax=\toX+0.45,
    xtick={\logsh-1,\logsh,\logsh+1}, xticklabels={$10^{-1}$,$10^{0}$,$10^{1}$},
    yticklabels={Prop I-3, Prop I-6, Prop I-12, Prop I-18, Prop I-20,
                 Prop I-27, Prop I-30, Prop I-36, Prop I-45,
                 Prop II-3, Prop II-12,
                 Prop III-6, Prop III-11, Prop III-25, Prop III-36},
    clip=false,
]
    \addplot[forget plot, bar shift=4.5pt, barbase]  table[x expr=ln(\thisrow{base_r1})/ln(10)+\logsh, y=ypos] {\wall};
    \addplot[forget plot, bar shift=3.0pt, barbase]  table[x expr=ln(\thisrow{base_r2})/ln(10)+\logsh, y=ypos] {\wall};
    \addplot[forget plot, bar shift=1.5pt, barbase]  table[x expr=ln(\thisrow{base_r3})/ln(10)+\logsh, y=ypos] {\wall};
    \addplot[forget plot, bar shift=-1.5pt, barmine] table[x expr=ln(\thisrow{mine_r1})/ln(10)+\logsh, y=ypos] {\wall};
    \addplot[forget plot, bar shift=-3.0pt, barmine] table[x expr=ln(\thisrow{mine_r2})/ln(10)+\logsh, y=ypos] {\wall};
    \addplot[forget plot, bar shift=-4.5pt, barmine] table[x expr=ln(\thisrow{mine_r3})/ln(10)+\logsh, y=ypos] {\wall};
    \draw[dashed, black!55, line width=0.6pt] ({axis cs:\toX,0}|-{rel axis cs:0,0})
        -- ({axis cs:\toX,0}|-{rel axis cs:0,1});
    \node[rotate=-90, anchor=south west, font=\scriptsize, black!55, inner sep=2pt]
        at ({axis cs:\toX,0}|-{rel axis cs:0,1}) {Timeout (12h)};
    \addplot[forget plot, only marks, mark=x, mark size=2.6pt, line width=0.9pt,
             cBaselineD]
      coordinates {(\toX+0.18,3.78)(\toX+0.18,5.78)(\toX+0.18,6.78)(\toX+0.18,7.78)
                   (\toX+0.18,8.78)(\toX+0.18,9.78)(\toX+0.18,11.78)
                   (\toX+0.18,12.78)(\toX+0.18,13.78)};
\end{axis}
\end{tikzpicture}
\caption{Synthesis wall-clock (hr)}
\label{fig:ablation-wall}
\end{subfigure}%
\hfill
\begin{subfigure}[t]{\compfrac\linewidth}
\centering
\begin{tikzpicture}
\begin{axis}[rq2axis,
    width=1.3\linewidth,
    xmode=log, log basis x=10,
    xmin=4, xmax=60,
    yticklabels={},
]
    \def\crosspos{4.6}
    \addplot[forget plot, bar shift=4.5pt, barbase]  table[x=base_r1, y=ypos] {\comp};
    \addplot[forget plot, bar shift=3.0pt, barbase]  table[x=base_r2, y=ypos] {\comp};
    \addplot[forget plot, bar shift=1.5pt, barbase]  table[x=base_r3, y=ypos] {\comp};
    \addplot[forget plot, bar shift=-1.5pt, barmine] table[x=mine_r1, y=ypos] {\comp};
    \addplot[forget plot, bar shift=-3.0pt, barmine] table[x=mine_r2, y=ypos] {\comp};
    \addplot[forget plot, bar shift=-4.5pt, barmine] table[x=mine_r3, y=ypos] {\comp};
    \addplot[forget plot, only marks, mark=x, mark size=2.8pt, line width=0.9pt,
             cBaselineD]
      coordinates {(\crosspos,3.78)(\crosspos,5.78)(\crosspos,6.78)(\crosspos,7.78)(\crosspos,8.78)
                   (\crosspos,9.78)(\crosspos,11.78)(\crosspos,12.78)(\crosspos,13.78)};
\end{axis}
\end{tikzpicture}
\caption{Compile time (s)}
\label{fig:ablation-compile}
\end{subfigure}
\caption{
    Fill Stage ablation over Books I--III (all $3$ runs shown as separate bars).
    (\subref{fig:ablation-wall})~Synthesis wall-clock per proposition; baseline runs that reach the $12$h cap fail, while every run of \methodname finishes within ${\sim}3.5$h.
    (\subref{fig:ablation-compile})~Compile time of the proof each arm \emph{produced}, timed by recompiling only that proof's own files from a warm cited-dependency cache.
}
\label{fig:ablation}
\end{figure}

%% file: figures/fig-compile-cactus.tex
\colorlet{cCactusOurs}{cMine}     
\colorlet{cCactusBase}{cBaseline} 
\begin{figure}[t!]
\centering
\begin{tikzpicture}
\begin{axis}[
    width=0.46\textwidth, height=5cm,
    ymode=log,
    ymin=3, ymax=6000,
    xmin=0, xmax=49,
    xlabel={Number of Propositions Compiled},
    ylabel={Compile Time (s)},
    xlabel style={font=\footnotesize}, ylabel style={font=\footnotesize},
    tick label style={font=\scriptsize},
    grid=major, grid style={black!8},
    legend style={font=\scriptsize, at={(0.03,0.97)}, anchor=north west, draw=black!45, fill=white},
    legend cell align=left,
    unbounded coords=discard,
    cactusrun/.style={semithick, mark size=1.5pt, mark repeat=5},
]
    \addplot[cactusrun, cCactusOurs, mark=*,        mark phase=0]
      table[x=rank, y=new_r1] {data/compile/cactus_runs.dat};
    \addlegendentry{\methodname (Ours) -- Run 1}
    \addplot[cactusrun, cCactusOurs, mark=triangle*, mark phase=2]
      table[x=rank, y=new_r2] {data/compile/cactus_runs.dat};
    \addlegendentry{\methodname (Ours) -- Run 2}
    \addplot[cactusrun, cCactusOurs, mark=square*,   mark phase=4]
      table[x=rank, y=new_r3] {data/compile/cactus_runs.dat};
    \addlegendentry{\methodname (Ours) -- Run 3}
    \addplot[cactusrun, cCactusBase, mark=*,        mark phase=0]
      table[x=rank, y=old_r1] {data/compile/cactus_runs.dat};
    \addlegendentry{Baseline (LeanEuclid) -- Run 1}
    \addplot[cactusrun, cCactusBase, mark=triangle*, mark phase=2]
      table[x=rank, y=old_r2] {data/compile/cactus_runs.dat};
    \addlegendentry{Baseline (LeanEuclid) -- Run 2}
    \addplot[cactusrun, cCactusBase, mark=square*,   mark phase=4]
      table[x=rank, y=old_r3] {data/compile/cactus_runs.dat};
    \addlegendentry{Baseline (LeanEuclid) -- Run 3}
    \draw[dashed, cCactusBase, line width=0.8pt, -{Latex[length=1.6mm]}]
      (axis cs:41,415) -- (axis cs:41,4600);
    \node[font=\scriptsize, cCactusBase, anchor=south] at (axis cs:41,4600) {?};
    \draw[dashed, black!45] (axis cs:0,3600) -- (axis cs:49,3600);
    \node[font=\scriptsize, black!55, anchor=north east] at (axis cs:49,3600) {Timeout (1 Hour)};
\end{axis}
\end{tikzpicture}
\caption{
    Cactus plot of cold-compile time on a log scale, all $3$ runs drawn per world.
    Our decomposed proofs compile all $48$ propositions, while the original baseline compiles $41$.
}
\label{fig:compile-cactus}
\end{figure}
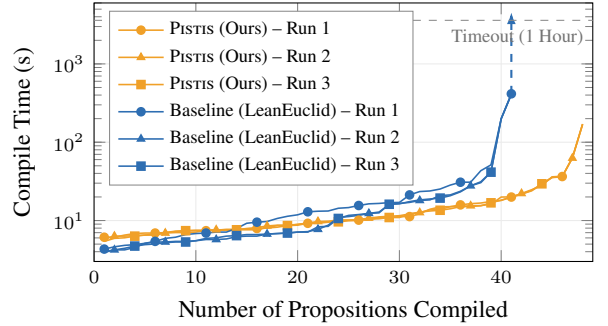

%% file: figures/fig-citation-mistake-img.tex
\begin{figure}[t]
\centering
\includegraphics{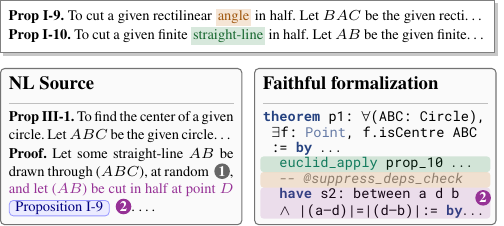}
\caption{
  A citation gap unveiled in the proof of Prop. III-1.
  Fitzpatrick brackets ``cut $AB$ in half'' as \refpill{Proposition I-9}, but
  I-9 is about \emph{angles}, not straight lines, and it should be Prop~I-10.
}
\label{fig:citation-mistake}
\end{figure}

%% file: figures/fig-accept-refute-img.tex
\begin{figure}[t]
\centering
\includegraphics{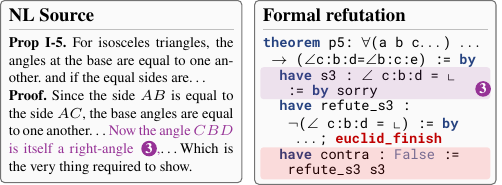}
\caption{
  Formal refutation of an LLM-generated proof.
  \textit{Left:} excerpt from an LLM-generated proof with false claim $s_3$.
  \textit{Right:} we prove $\neg\formal{\assert{s_3}}$.
}
\label{fig:accept-refute}
\end{figure}

%% file: sections/6-related.tex
\section{Related Work}
\label{sec:related}

\paragraph{Autoformalization and Euclidean Geometry.}
Autoformalization in Lean~\cite{demoura2021lean} grew from landmark formalizations and libraries~\cite{gonthier2008four,hales2017formal,mathlib2020} into an area driven by language models~\cite{wu2022autoformalization,li2024survey}.
Most work asks what systems \emph{can} formalize, at growing scales: competition and textbook statements~\cite{zheng2022minif2f,azerbayev2023proofnet}, repositories and long contexts~\cite{yang2023leandojo,hu2024minictx}, corpora and theories~\cite{min2026divide,feng2026lcsbench}, progress measured by compilation.
Attention has recently turned to \emph{how} a proof closes.
Euclid has drawn the most: Avigad et al.~\cite{avigad2009formal} give system $E$ as a \emph{faithful} model of his proofs, Beeson et al.~\cite{beeson2019proof} machine-check Book~I, and LeanEuclid~\cite{murphy_autoformalizing_2024} implements $E$ in Lean with SMT for diagrams.
Elsewhere informal sketches guide proof search~\cite{jiang2023draft,zheng2024lyra}, but only heuristically.

\paragraph{Neurosymbolic and Agentic Approaches.}
Neurosymbolic programming~\cite{huang2021scallop,li2023scallop,manhaeve2018deepproblog} supplies neural facts to declarative programs reasoned over symbolically; agentic coding frameworks~\cite{wang2025openhands} establish the tool-equipped-agent pattern.
Closest are hierarchical decomposition searches~\cite{lample2022hypertree,goedelcodeprover}, which split a goal into independently closed subgoals chosen for \emph{tractability}: any that closes the theorem will do, none constrained by the source argument.

\paragraph{Faithfulness.}
Statement-level faithfulness means semantic equivalence with the source, via type-checking, back-translation, or logical equivalence~\cite{poiroux2025reliable,lu2024formalalign}.
Such equivalence never constrains the proof: the kernel accepts any well-typed term inhabiting the goal, so a faithful statement admits proofs the source never gave.
Proof-level evaluation is weaker: LeanEuclid~\cite{murphy_autoformalizing_2024} measures only whether Lean accepts the proof and its edit distance to a hand repair.
ProofFlow~\cite{proofflow} preserves a dependency graph but validates only well-formedness, not whether each step's tactics use its declared premises.
The informal analogue asks whether a chain of thought reflects its computation~\cite{turpin2023language}.

%% file: sections/7-conclusion.tex
\section{Conclusion and Future Works}
\label{sec:conclusion}

We introduced \methodname, an oracle-guided proof search that formalizes a natural-language proof into a Lean proof satisfying \faithfulcriteria.
Across three books of Euclid's \emph{Elements}, \methodname produces proofs that are more faithful than prior work and compile faster, and it accepts or refutes human- and AI-written arguments, surfacing genuine errors.
We plan to extend faithful formalization beyond Euclidean geometry and reduce its reliance on supervision.

\paragraph{Limitations.}
Our study is limited to Euclidean geometry in System~E, which cannot express a handful of \emph{Elements} propositions.
Moreover, \methodname relies on a human or LLM oracle to certify faithfulness, and our conditions, while necessary, are not sufficient to capture an author's full intent.

%% file: sections/8-appendix.tex
\appendix


\section{\methodname Implementation Details}
\label{app:pistis-impl}

This section is intended to clarify the connection between what the main paper describes abstractly in the methodology to how it is implemented. Examples of such questions we answer include: what does the mapping $\formal{\cdot}$ look like in the code? How do we enforce that the agent follows \fillalgoname?

\subsection{How Objects are Implemented}
\label{app:objects}

\paragraph{Input Objects.}

\input{figures/fig-input-prop6}

Recall (\cref{sec:methodology}) that \methodname takes as input the tuple
$\langle \propN, \formal{\propN}, \PropNProof, \systemEAxioms \rangle$.
\cref{fig:input-prop6} instantiates every component \emph{verbatim from the
released artifact} on Book~I, Prop.~6 (I-6), the running example of
\cref{fig:motivation}, so a reader can locate each object in the
codebase directly.
The figure shows every input object, as well as further details such as what the free variables are, and how $\formal{\propN}$ is broken down into $\hypothesis \to \conclusion$.

\paragraph{Free Variables.}
\label{app:why-free-vars}
We explain why there are free variables rather than having the variables quantified with $\forall$.

Usually, theorems never have any free variables, and they are all quantified with either $\forall$ or $\exists$. However the way we write theorems in Lean, when we \emph{prove} a theorem, we are not proving the one with $\forall$ quantifiers, and rather the one with free variables.
Take for example the theorem statement $\forall x \in C, P(x) \to Q(x)$.
In Lean this theorem statement is
\[ \texttt{theorem prop : (x : C), P(x) \(\to\) Q(x) :=} \]
This is a \emph{function}, where $x \in C$ is \emph{already given and fixed}. The proof, that is the stuff after :=, is really a proof that $\systemEAxioms \cup \{x \in C\} \vdash P(x) \to Q(x)$. Here, $x$ is a \emph{free} variable \emph{not} quantified with $\forall$. Another subtlety, is in this case, we would consider $P(x)$ as the hypothesis, and because we use the \texttt{intro} tactic, we are really proving $\systemEAxioms \cup \{x \in C, P(x)\} \vdash Q(x)$. Hence in this example, $\hypothesis=\alpha_1 \land \alpha_2 = x \in C \land P(x)$, while $\conclusion=Q(x)$, with the proof being that of $\systemEAxioms \cup \{x \in C, P(x)\} \vdash Q(x)$.

As a final note, even though Lean only directly proves $\systemEAxioms \cup \{x \in C, P(x)\} \vdash Q(x)$, this directly implies  $\systemEAxioms \vdash x \in C \to (P(x) \to Q(x))$, which further implies $\systemEAxioms \vdash \forall x \in C, (P(x) \to Q(x))$

\paragraph{Map Phase Objects.}
The \mapstage produces the tuple $M=\langle \bar{\phi'}, \bar{s}, \assumptions{\cdot},
\assert{\cdot}, \formal{\cdot}\rangle$ (\cref{sec:methodology-overview}) as a single Lean file,
and \cref{fig:map-prop6} shows this artifact concretely for Prop.~6 in its \emph{pre-fill} state,
where every step body is still \texttt{:= by sorry}.
The \texttt{euclid\_sentence} tactic is what constructs these objects, and each output-tuple
component is pinpointed in the figure. Each tuple object is
implemented as:
\begin{itemize}[leftmargin=1.4em,itemsep=1pt,topsep=2pt]
\item $\bar{s}=\{s_1,\dots,s_n\}$, the ordered \NL string literals, one per
\texttt{euclid\_sentence}, appearing top-to-bottom so the \order condition holds by construction;
\item $\formal{\assert{s_i}}$, the \emph{type} annotation $\tau$ of $(\texttt{step}i : \tau)$ in each
sentence. Also, the assertion $\assert{s_i}$ is the part of $s_i$ that is \emph{not} an assumption;
\item For each $a \in \assumptions{s_i}$, $a$ and $\formal{a}$ are defined by the \texttt{-{}- @assumption
("\textit{text}", \textit{type})} comment tags above the sentence, one per assumption $a$;
\item $\bar{\phi'}$, the sequence of subgoals the \fillstage will close, which consists of
\emph{both} the sentence steps $\formal{\assert{s_i}}$ \emph{and}
the additional helper \texttt{have \dots := by sorry} steps the map stage inserts to bridge them
(e.g. \texttt{swapfig} in \cref{fig:map-prop6}).
\end{itemize}
Alongside \texttt{euclid\_sentence}, three sibling tactics handle Euclid's non-assertive sentences.
These still record the text (so \coverage can tile the full proof) but emits no \texttt{have}, carries no claim, and leaves the proof term and context unchanged. The three differ only by what role they mark:
\begin{itemize}[leftmargin=1.4em,itemsep=1pt,topsep=2pt]
\item \texttt{euclid\_intro\_sentence}: this is just $\propN$
\item \texttt{euclid\_conclude\_sentence}: the closing restatement of $\propN$, which Euclid always writes
\item \texttt{euclid\_wts}: a \emph{mid-proof} ``I say that~$X$'' (last line of \cref{fig:map-prop6})
that announces the goal $X$ the following sentences will establish, rather than asserting $X$ is
already proven.
\end{itemize}
Because a structural sentence asserts nothing, for it $\assert{s_i}$ is the empty string,
$\assumptions{s_i}=\varnothing$, and $\formal{\assert{s_i}}=\texttt{True}$.

\input{figures/fig-map-prop6}

\paragraph{Fill Stage Objects.}
The \fillstage proves each $\phi'_i$ that the map stage left as \texttt{sorry}: for each it produces a
sequence of steps $\kappa_1, \dots, \kappa_t$ such that Lean checks
$\systemEAxioms \cup \{\hypothesis, \phi'_1, \dots, \phi'_{i-1}, \kappa_1, \dots, \kappa_t\} \vdash \phi'_i$
(\cref{sec:methodology-overview}).
\cref{fig:fill-prop6} shows an example of how this happens concretely.

\input{figures/fig-fill-prop6}

\input{figures/fig-birdseye-prop6}
\paragraph{Output Objects.}
\cref{fig:birdseye-prop6} gives a bird's-eye view of
a fully completed proposition, showing an example of what the full output looks like in practice. 

\subsection{Map Stage Details}
This section explains how the map stage is executed in the code.
\label{app:map-stage}

The map stage runs as \emph{skills} (\cref{tab:skills}) which are prompt files. The first skill ran is
\textbf{(1) \texttt{faithful-split}}, which gets the LLM to read the proposition's English text and splits it into an ordered list of NL segments satisfying \atomicity and \coverage, and finally written to a file \texttt{split.json}. 
A \script (\texttt{check\_faithful.py -{}-split}) then confirms this partition tiles the source text byte-for-byte, enforcing \coverage before any Lean is written. 
Then \textbf{(2) \texttt{faithful-map}} turns that split into \texttt{Main.lean}.
It first stamps a placeholder file, a \texttt{euclid\_sentence "Book\#.Prop\#.Step\#" "NL segment" (stepN : True) :=
by sorry} per entry, using the \texttt{faithful\_map\_assemble.py -{}-placeholders} script, which
copies each sentence's \texttt{"NL segment"} and locator directly from \texttt{split.json}.
Then the agent replaces each \texttt{True} with (what the agent thinks) is the sentence's real claim $\formal{\assert{s_i}}$ and each
placeholder tag with a real \texttt{@assumption ("text", type)}.

At the end \texttt{check\_faithful.py} re-checks \coverage and \order, while the oracle checks \atomicity and \atomicfaithfulness.

One may wonder how \order is possible to check. After all, the definition in \cref{tab:conditions} references $\bar{\phi}$ the \emph{final} proof, which does not yet exist.
Recall that the \emph{fill stage} is only allowed to insert formulae $\kappa$ \emph{in between} two existing $\phi'_i, \phi'_j$. This is enforced in practice by telling the LLM, and script checkers such as \texttt{check\_faithful.py}. Thus checking the order is correct in the map stage guarantees the order stays correct in the fill stage.

Once the \oracle approves all checks in the end, \texttt{check\_steps.py -{}-save} saves these objects so the fill stage cannot silently alter them (the LLM agent is denied write access).

\subsection{Fill Stage (\fillalgoname) Details}
\label{app:fillalgo-details}

Similar to how map stage is implemented, the fill stage relies on skills, but also relies on hooks, and tools to follow \fillalgoname.

It starts by calling the \texttt{faithful-prove} skill (\cref{tab:skills}), which contains the core instructions of \fillalgoname. During the algorithm, a set of tools (\cref{tab:agent-tools}) help the agent in the proof search process, and hooks/settings make the agent actually follow the algorithm rather than merely being told to. 
One example of a setting to enforce good behavior, is that the agent is denied raw \texttt{lake build}, and can only build using one of the tools, so all its build commands can be checked.

An example of a hook, is that we hard-deny the agent from writing any
\texttt{stepN.lean} that is past the latest step checked, so it physically cannot skip
ahead or prove a later step before an earlier one is certified, enforcing the in-order iteration of
line~1 of \cref{alg:orderdecompose}. Routing every build through these scripts likewise forces the
SF/SP/P (sufficiency, suppliability, and provability) discipline on each new lemma: it is first checked for sufficiency (line~16), then suppliability
(line~18), and only then proven (line~20). A final \texttt{check\_step.py -{}-all} certifies the whole
proof end-to-end; in practice it rarely surfaces anything new, since each step was already certified in
isolation as it was proved.

\paragraph{Resume.}
\texttt{check\_step.py} records which steps are already proven. Because the agent proves the steps in
order, if it stops after finishing $\phi'_{i-1}$ it simply continues from $\phi'_i$ rather than starting
over. This is correct because each $\phi'_i$ is proved using only $\hypothesis, \phi'_1, \dots,
\phi'_{i-1}$, so it can never depend on a later step; the steps already proven stay valid, and if the
\oracle later fixes a map-stage mistake only that step and the ones after it need re-proving.

\begin{table*}[t]
\centering
\small
\begin{tabular}{lp{0.30\textwidth}p{0.42\textwidth}}
\toprule
\textbf{Skill} & \textbf{Role} & \textbf{Example instruction (verbatim excerpt)} \\
\midrule
\texttt{faithful-split} & Map stage, part~1: split the \NL proof into atomic assertions & ``Your ONLY job: take the raw English text of a Euclid proposition and split it into atomic assertions. [\dots] You do NOT translate to Lean, you do NOT look at any \texttt{.lean} file.'' \\
\addlinespace
\texttt{faithful-map} & Map stage, part~2: turn each assertion into a Lean claim type & ``Your job: fill \dots a Lean claim type per sentence \dots one sentence at a time. [\dots] RULE~0: THE SENTENCE IS THE CLAIM. [\dots]'' \\
\addlinespace
\texttt{faithful-prove} & Fill stage: main instructions to follow \fillalgoname & ``prove each Euclid sentence's step, IN ISOLATION, using the recursive... \dots the SCRIPT does all wiring \dots you only ever write proof bodies and add \texttt{have}+backing-file decompositions.'' \\
\bottomrule
\end{tabular}
\caption{The agent skills (prompt files under \texttt{.claude/skills/}) and a verbatim excerpt of each one's core instruction. The map stage runs \texttt{faithful-split} then \texttt{faithful-map}; the fill stage then runs \texttt{faithful-prove}.}
\label{tab:skills}
\end{table*}

\begin{table*}[t]
\centering
\small
\begin{tabular}{lp{0.68\textwidth}}
\toprule
\textbf{Tool} & \textbf{Purpose and Role in Faithfulness} \\
\midrule
\texttt{check\_step.py} & The agent's only build tool in Phase B. Implements SF/SP/P checks for each $\phi'_i$. \\
\addlinespace
\texttt{check\_step -{}-context} & Dumps exact hypotheses available at a proof node, showing context $\systemEAxioms \cup \{\phi'_1, \dots, \phi'_{i-1}\}$ for proving $\phi'_i$. Operationalizes \textsc{InContext} predicate (\cref{alg:orderdecompose}, line 5). \\
\addlinespace
\texttt{find.py} & Declarative query interface over indexed database of System~E axioms, definitions, and previously proven propositions. Agent uses it during \textsc{CreateLemmas} (\cref{alg:orderdecompose}) to locate relevant axioms, or propositions. An example is that agents can search for all formal statements that conclude a specific property. \\
\addlinespace
\texttt{scaffold\_step.py} & Generates boilerplate backing files for each $\phi'_i$ left as \texttt{sorry} after mapping stage.\\
\bottomrule
\end{tabular}
\caption{Examples of custom Python tools provided to LLM agents during the filling stage. }
\label{tab:agent-tools}
\end{table*}


\section{Accept, Refute, and Gaps}
\label{app:accept-refute-gaps}
In this section, we validate our formal refutation methods, then go into more detail on gaps.

\subsection{Why Refutation Creates a Logical Contradiction}
\label{app:refutation-false}
This section goes into detail how we are able to prove a logical contradiction if either of the following are true:
\begin{enumerate}
    \item[A] $\exists i, \systemEAxioms \cup \{\hypothesis\} \vdash \neg \formal{\assert{s_i}}$
    \item[B] $\exists i, \systemEAxioms \vdash \exists v_1, ..., v_k, \neg (\hypothesis \to \formal{\assert{s_i}})$, where $v_1, ..., v_k$ are all the free variables occurring in $\alpha$
\end{enumerate}
We begin by assuming that System E axioms cannot prove the contradiction, denoted $\texttt{False}$ in Lean, and nor can the hypotheses ( $\systemEAxioms \cup \{\hypothesis\} \nvdash \texttt{False}$).
Recall that \emph{if}, \methodname is able to accept the proof, then a proof $(\phi_1, ...., \phi_N)$ of $\systemEAxioms \cup \{\hypothesis \} \vdash \conclusion$, is outputted.
A property of this proof, is that for every $\phi_i$, $\systemEAxioms \cup \{\hypothesis \} \vdash \phi_i$ (according to first order logic). Also, recall that every $\formal{\assert{s_i}}=\phi_{k}$ for some $k$ (output 4 of \cref{sec:methodology}).

If A is true, then $\systemEAxioms \cup \{\hypothesis\} \vdash \neg \phi_k$, but because $\bar{\phi}$ is a proof of $\systemEAxioms \cup \{\hypothesis\} \vdash \conclusion$, $\systemEAxioms \cup \{\hypothesis\} \vdash \phi_k$ hence we prove \texttt{False}. We assumed we cannot prove \texttt{False} so this is a contradiction.

If B is true, we show that it will also lead to a contradiction. Like case A, $\systemEAxioms \cup \{\hypothesis\} \vdash \phi_k$, and therefore $\systemEAxioms \vdash \hypothesis \to \phi_k$, which implies that for variables $v_1, ..., v_k$ occurring free in $\hypothesis$: $\systemEAxioms \vdash \forall v_1, ..., v_k, (\hypothesis \to \phi_k)$.
But by assumption B, $\systemEAxioms \vdash \exists v_1, ..., v_k, \neg (\hypothesis \to \phi_k)$, so  $\systemEAxioms \vdash \neg (\forall v_1, ..., v_k, (\hypothesis \to \phi_k))$ hence we prove \texttt{False}.

\subsection{Assumption Gap Implementation Details}
\label{sec:assumption-tagging}
Recall the (optional) assumption gap tagging stage, which tries automated tactics to close the assumption. This section goes into more detail, such as explaining the exact tactics to close the assumption. 

The exact automated tactics tried are: \texttt{rfl}, \texttt{assumption}, \texttt{simp}, \texttt{linarith}, \texttt{nlinarith}, \texttt{euclid\_finish}, where \texttt{euclid\_finish} is essentially an SMT solver.
These are tried one at a time in the order listed, and  if none of them close the assumption, we mark it as a gap (\texttt{@assumption\_gap}) and leave a \texttt{sorry} placeholder.

We justify this choice of measuring gaps since assumptions are notions that \emph{should already be proven}, not new steps to prove, and we believe these automated tactics are generous (they are already powerful), so if they claim there is an assumption gap, then it is trustworthy.

Across our \numproofs proofs, the stage materializes \numAssumpTags assumptions in total, of which \numAssumpGapTags are marked as gaps and passed to the \fillstage (\numAssumpGapTagsBookI, \numAssumpGapTagsBookII, and \numAssumpGapTagsBookIII{} in Books~I, II, and III respectively).

\subsection{Citation Gap}
\label{app:citation-gap}

A \emph{citation gap} is a cited \texttt{[Prop.~B.N]} that the \citcond check cannot satisfy. We mark it
with a \texttt{-{}- @suppress\_deps\_check "reason"} line above the sentence, which tells the checker script 
(\texttt{check\_faithful.py}) to skip that one citation. This is needed in two cases: the citation is an
actual mistake in the source edition (as explained in the main text), or it points to a proposition we
skipped (one of the Book~III propositions we leave out of scope; see \cref{app:limitations}).
We use \numSuppressCiteTags{} such tags across Book~III.

\subsection{Generic Gap Implementation Details}
\label{app:generic-gap}

A \emph{generic gap} is a gap in Euclid's reasoning itself, distinct from an assumption and citation gap.
These surface when \fillalgoname cannot close a step and returns \texttt{False}: if the \oracle agrees the
step is a genuine gap rather than an \llm or mapping failure, we mark it with a
\texttt{-{}- @euclid\_gap: <why>} comment.

For example, Book~III Prop.~35 says that if two straight-lines $AC$ and $BD$ in a circle cut one another
at $E$, the rectangle on the two pieces of $AC$ equals the rectangle on the two pieces of $BD$. Euclid
proves two cases: both lines pass through the centre, and neither line passes through the centre. He does
not treat the remaining case, where one line passes through the centre and the other does not, even
though his hypotheses allow it. We add this missing case and tag the site
\texttt{-{}- @euclid\_gap}.
Across Book~III we mark \numTotalEuclidGaps{} such generic gaps. None were marked in Book~II and we do not claim any gaps in Book~I as these were already found by \cite{murphy_autoformalizing_2024}.

Note ordinary System-E
plumbing should not be considered a gap, and what truly constitutes a gap is subjective so these are simply what we judged to be a gap.

\section{Human Evaluation Protocol}
\label{app:human-eval-rubric}

Reviewers used a custom web interface to evaluate anonymized pairs of
formalizations. The interface supported the full review workflow, from an
interactive tutorial to step-fidelity scoring and final pairwise preferences.
The review itself proceeded in two stages.
\begin{enumerate}[leftmargin=*,itemsep=1pt,topsep=2pt]
    \item They rated the step fidelity of each formalization independently.
    \item They answered pairwise preference questions comparing the two
    anonymized formalizations.
\end{enumerate}

To ensure fairness, \emph{we compare only the formal Lean proofs}, and \emph{remove any non-essential objects}. That is, out of the output objects $\langle \bar{\phi}, \bar{s}, \assumptions{\cdot}, \assert{\cdot}, \formal{\cdot} \rangle$ that \methodname creates, we remove $\bar{s}, \assumptions{\cdot}, \assert{\cdot}, \formal{\cdot}$ from the Lean file of our version, so we are only comparing the proof $\bar{\phi}$. We do this in practice by replacing $\texttt{euclid\_sentence}$ with $\texttt{have}$ and removing comments.

\subsection{Step Fidelity}
\label{app:step-fidelity}

The first rubric item reviewers were asked to rate is step fidelity. This was rated on a 0--5 scale for each formalization. Reviewers were
asked to assess how well the Lean proof preserves the mathematical route of the
textbook proof. The full scoring rubric is shown in
\cref{tab:step-fidelity-rubric}.

\begin{table}[tb]
\centering
\small
\begin{tabular}{cp{0.78\linewidth}}
\toprule
\textbf{Score} & \textbf{Description} \\
\midrule
0 & The Lean proof does not represent the textbook proof's mathematical content. \\
1 & Only isolated textbook-like facts appear. \\
2 & Some essential steps are present, but important objects, constructions,
dependencies, or conclusions are missing or hidden. \\
3 & The main route is recognizable, but there are notable omissions,
compression, ordering issues, or unclear automation. \\
4 & The essential steps and dependencies are represented with only minor formal
differences. \\
5 & The formalization closely preserves the textbook's mathematical route in a
readable and recoverable way. \\
\bottomrule
\end{tabular}
\caption{Step-fidelity rubric shown to reviewers.}
\label{tab:step-fidelity-rubric}
\end{table}

\subsection{Pairwise Preferences}
\label{app:pairwise-preferences}

For each proposition, reviewers answered three seven-tier pairwise preference
questions. The interface did not show numeric values: the center choice
indicated no preference, and choices farther to either side indicated stronger
preference for one anonymized method. \cref{tab:preference-rubric}
summarizes the three pairwise criteria.

\begin{table}[tb]
\centering
\small
\begin{tabular}{@{}p{0.32\linewidth}p{0.58\linewidth}@{}}
\toprule
\textbf{Metric} & \textbf{Question} \\
\midrule
Mathematical transparency & Which formalization makes the mathematical argument
more transparent and easier to follow? \\
Textbook representation & Which formalization better represents the textbook
proof? \\
Overall preference & Which formalization is preferred overall for the
proposition? \\
\bottomrule
\end{tabular}
\caption{Pairwise preference criteria used in the human evaluation.}
\label{tab:preference-rubric}
\end{table}

\subsection{Evaluation Interface}
\label{app:eval-interface}

\begin{figure*}[t]
\centering
\begin{subfigure}[t]{0.48\textwidth}
    \centering
    \includegraphics[width=\linewidth]{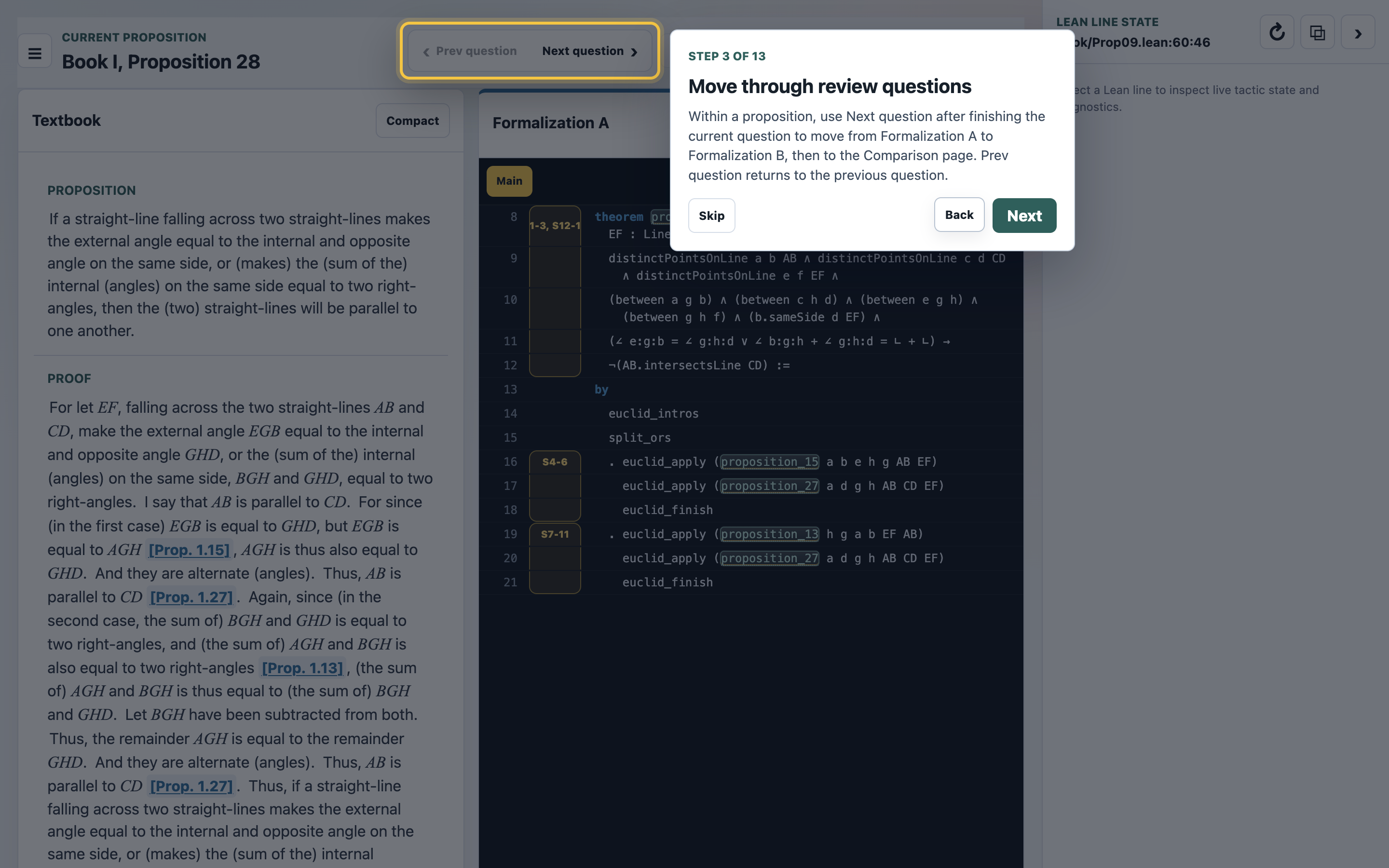}
    \caption{Tutorial}
\end{subfigure}\hfill
\begin{subfigure}[t]{0.48\textwidth}
    \centering
    \includegraphics[width=\linewidth]{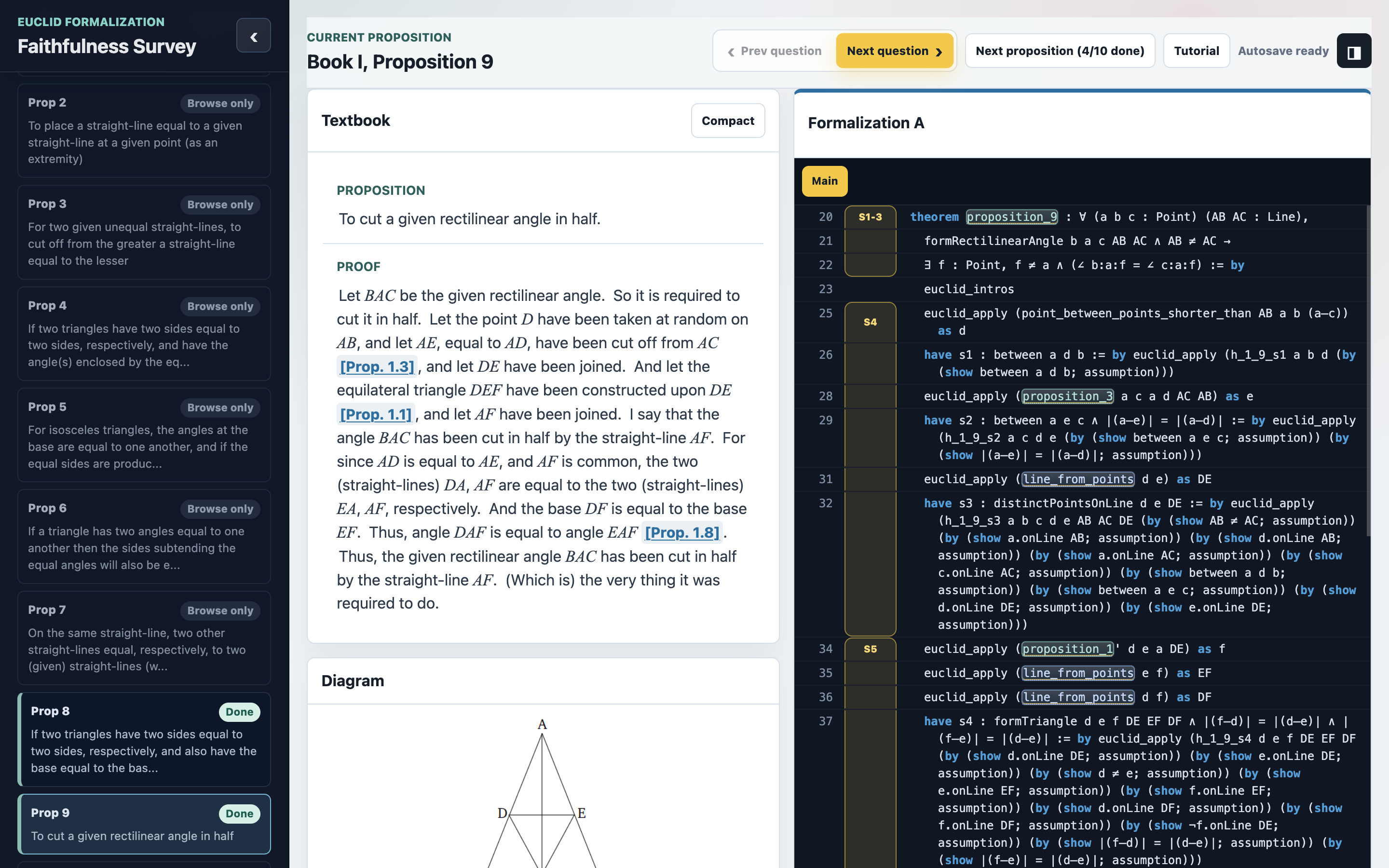}
    \caption{Textbook proof and formalization}
\end{subfigure}

\begin{subfigure}[t]{0.48\textwidth}
    \centering
    \includegraphics[width=\linewidth]{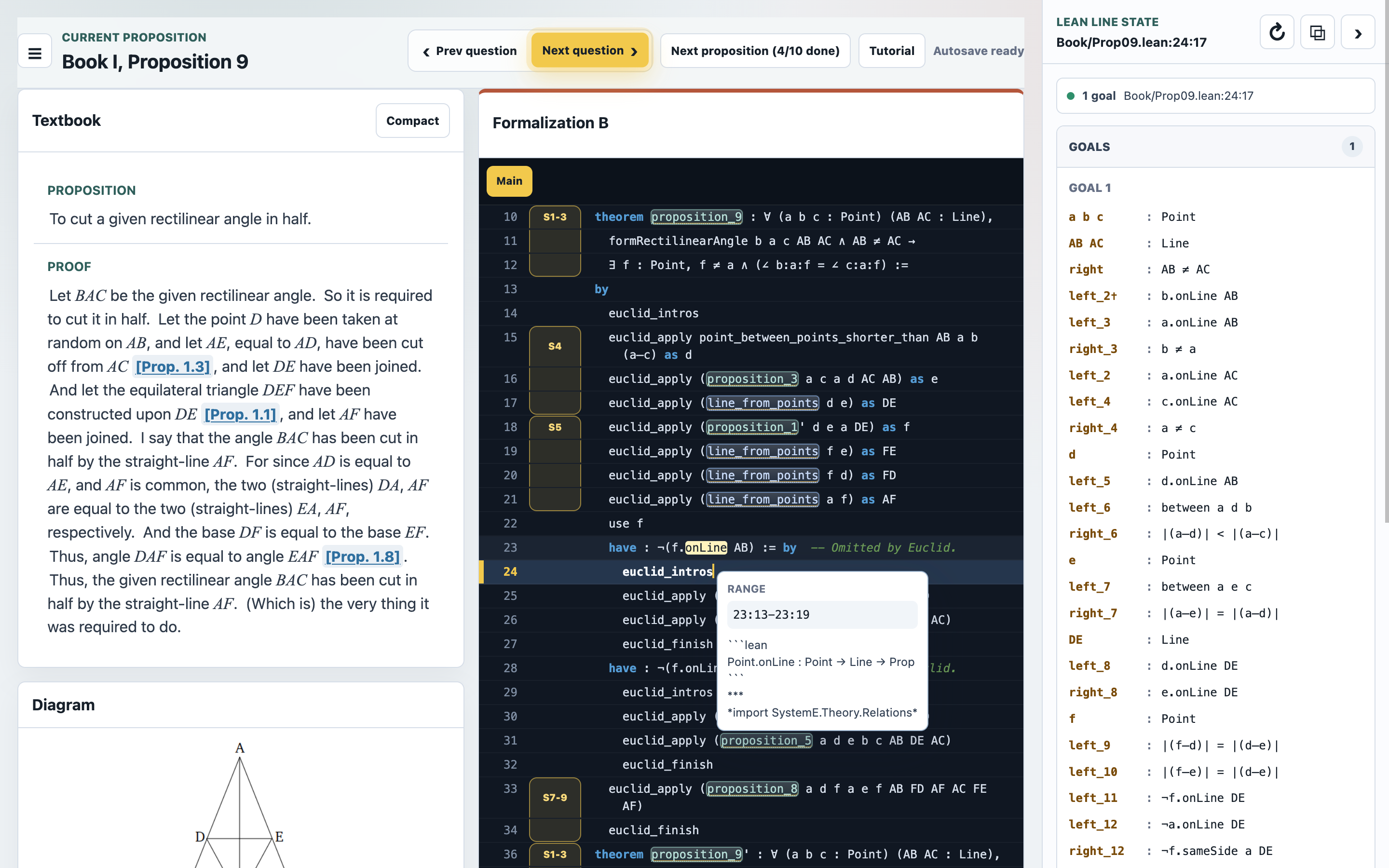}
    \caption{Live Lean query}
\end{subfigure}\hfill
\begin{subfigure}[t]{0.48\textwidth}
    \centering
    \includegraphics[width=\linewidth]{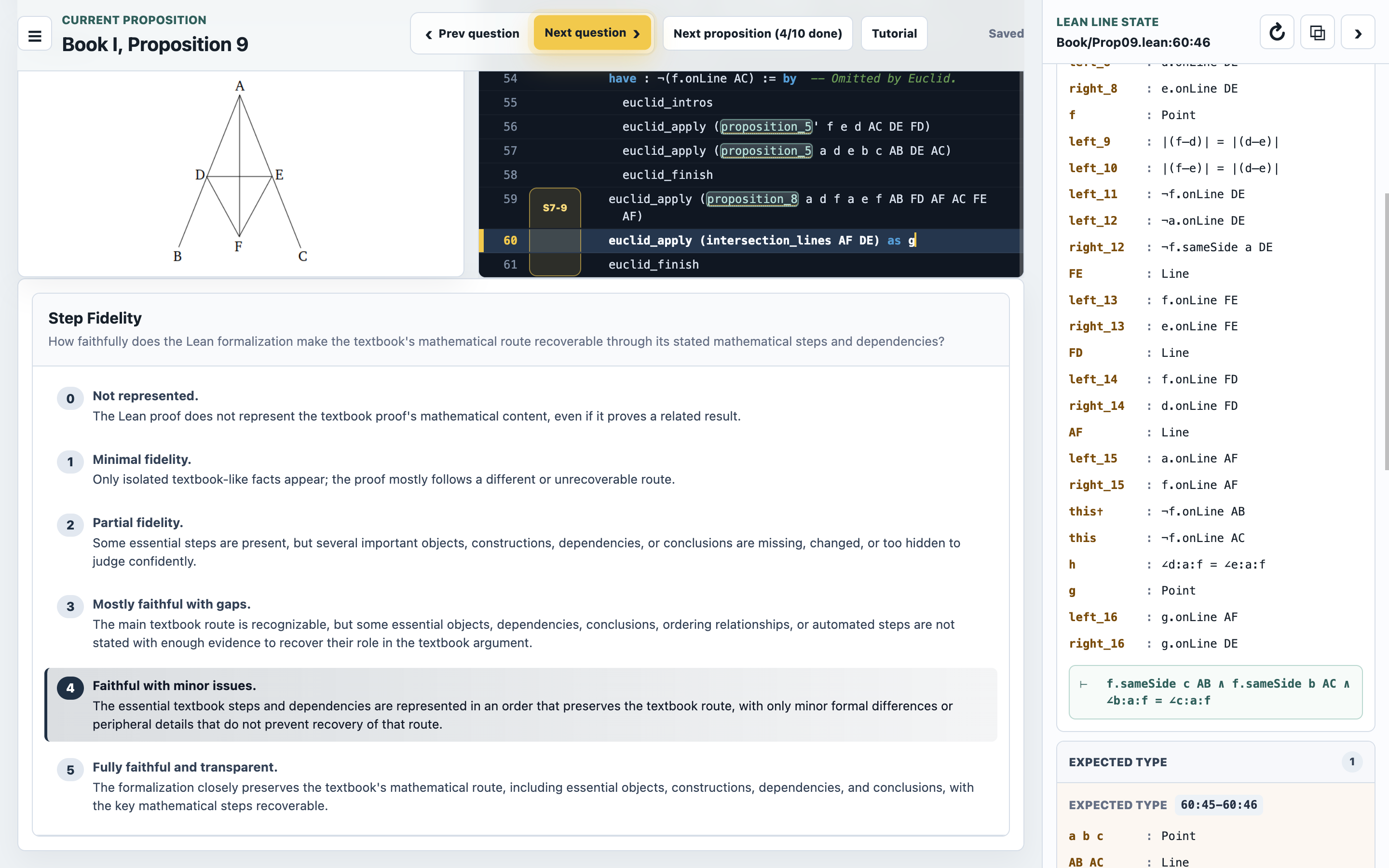}
    \caption{Step-fidelity scoring}
\end{subfigure}

\begin{subfigure}[t]{0.48\textwidth}
    \centering
    \includegraphics[width=\linewidth]{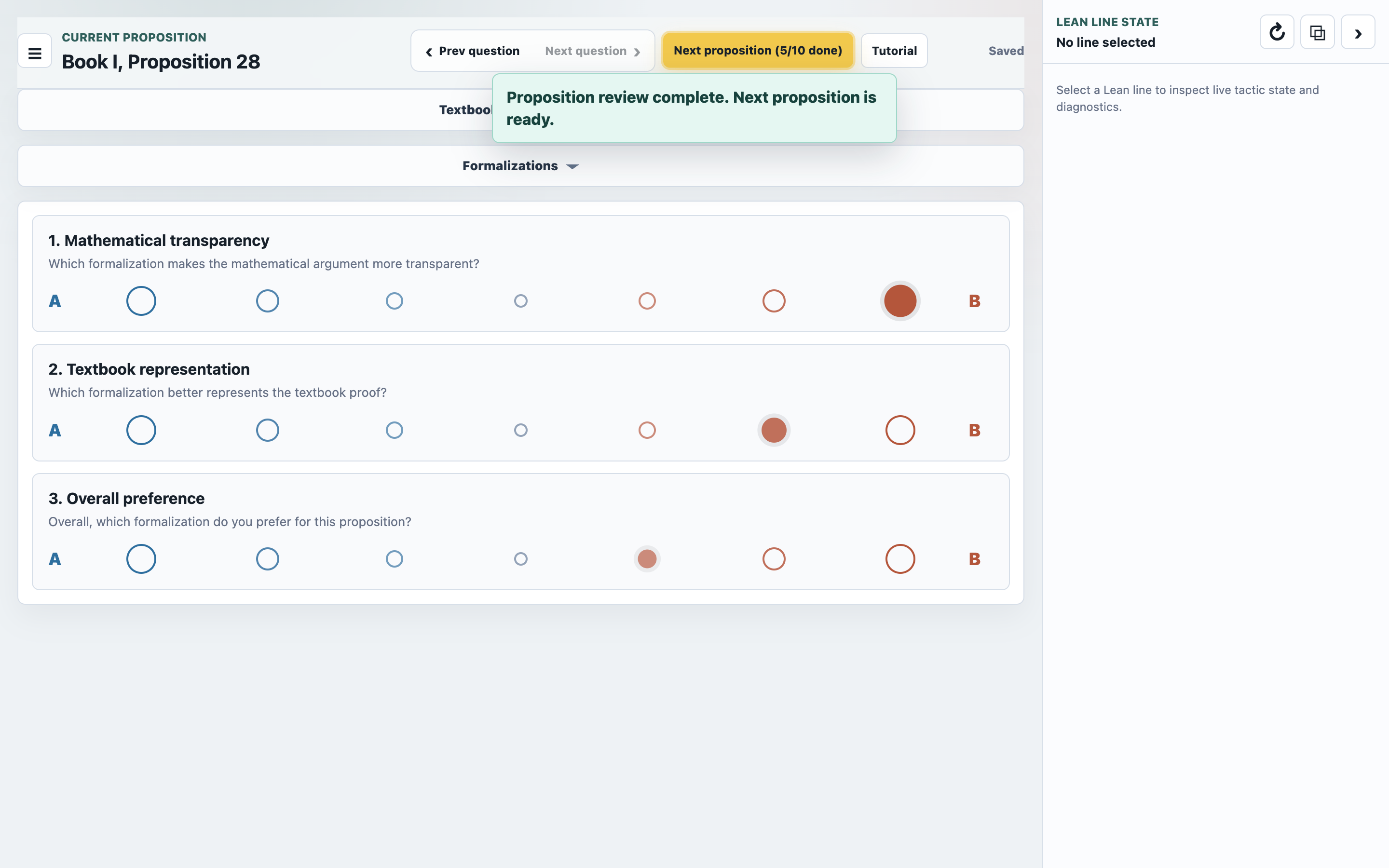}
    \caption{Pairwise preferences}
\end{subfigure}
\caption{Human-evaluation interface. Reviewers complete a tutorial, inspect each
anonymized formalization against the textbook proof, query a live Lean server,
score step fidelity with the rubric available in the interface, and finally
record pairwise preferences across the three preference criteria.}
\label{fig:survey-interface}
\end{figure*}

\Cref{fig:survey-interface} illustrates the website used for the
human-expert evaluation. Reviewers first completed an interactive tutorial that
introduced the task, the anonymized presentation, and the available controls.
For each proposition, the interface showed the textbook theorem and proof beside
one formalization at a time, so reviewers could compare the formal proof against
the natural-language argument without seeing the method identity. A live Lean
server was available for querying proof states and inspecting what each line
establishes. After examining each formalization, reviewers assigned a
step-fidelity score using the detailed 0--5 rubric shown in the interface, then
reported pairwise preferences for mathematical transparency, textbook
representation, and overall preference.

\section{LLM as a Judge}
\label{app:llm-judge}

Complementing the human study, we assess faithfulness with an LLM-as-judge protocol using two judges: a
closed model, Claude Opus~4.6 (whose radar is \cref{fig:llm-as-judge-radar} in the main text), and an
open model, GLM-5 (\cref{fig:llm-as-judge-radar-glm}). For each proposition the judge scores the formalization on five independent $0$--$5$ dimensions,
each defined by a strict rubric (\cref{tab:llm-rubric}): object correspondence, proof-step coverage,
proof-structure fidelity, cited-dependency fidelity, and assumption/side-condition fidelity.

\input{figures/fig-radar-glm}

The full prompt given to each judge, which embeds the shared five-dimension rubric of
\cref{tab:llm-rubric}, is:
\begin{lstlisting}[basicstyle=\ttfamily\footnotesize,breaklines=true,numbers=none,xleftmargin=1em]
You are an expert in Euclidean geometry, mathematical logic, and
formal verification in Lean 4.

Your task is to evaluate how faithfully a Lean formalization
represents the mathematical content of a textbook proposition and
proof from Euclid's Elements. The textbook is the reference standard.
You should not evaluate general elegance, maintainability, proof
brevity, annotation quality, or code style except insofar as these
affect mathematical faithfulness to the textbook.

Formal encoding overhead -- such as extra type parameters (e.g.
requiring a Line object where the textbook just names two points),
additional binders for type-checking, imported helper lemmas whose
internals are not shown, or sentence-level annotations -- is a routine
artifact of formalization in a proof assistant. Do not penalize a
formalization for these formal necessities. Evaluate only whether the
mathematical content (objects, claims, proof steps, dependencies,
assumptions) is faithful to the textbook.

You will score the formalization on 5 independent metrics, each on a
0-5 scale, using the rubric below.

[ RUBRIC -- the five-dimension scoring rubric ]

For each of the 5 metrics, provide a brief justification (2-4
sentences) citing specific parts of the Lean code and textbook proof.
After all justifications, output your final scores as five integers
(0-5) in angle brackets in the following order:
<object_correspondence, proof_step_coverage, proof_structure_fidelity,
 cited_dependency_fidelity, assumption_side_condition_fidelity>
For example: <4, 3, 5, 4, 3>
You MUST end your response with exactly one set of angle brackets
containing the five scores.

--- EVALUATION DATA ---
TEXTBOOK PROOF:      [ natural-language proof ]
LEAN FORMALIZATION (Main file):      [ Main.lean ]
HELPER LEMMA FILES (backing proofs for each sentence-level step):
     [ each stepN.lean, separated by --- ]
\end{lstlisting}

\begin{table*}[t]
\centering
\footnotesize
\begin{tabular}{@{}c p{0.90\textwidth}@{}}
\toprule
\textbf{Score} & \textbf{Description} \\
\midrule
\multicolumn{2}{@{}l}{\textbf{(a) Object Correspondence}} \\
\addlinespace[1pt]
0 & Object mapping invalid. The main objects are mapped incorrectly, making the formalization about different geometric entities than the textbook. \\
1 & Major object-role errors. Several important objects or roles are confused, omitted, or swapped in ways that materially change the proposition or proof. \\
2 & Partial object correspondence. The principal objects are identifiable, but important constructed objects, incidences, angle vertices, side correspondences, or role assignments are wrong or ambiguous. \\
3 & Mostly correct object correspondence with notable issues. Most objects are mapped correctly, but there are nontrivial local mismatches, such as an incorrect orientation, a missing constructed object, or a questionable side/angle correspondence. \\
4 & Faithful object correspondence with minor formal differences. All central objects and roles match the textbook. Minor discrepancies are limited to harmless naming, orientation conventions, or formal encodings that preserve the mathematical content. \\
5 & Fully faithful object correspondence. Every relevant textbook object, including constructed objects and their roles, is represented by the corresponding Lean object or relation with no substantive mismatch. \\
\midrule
\multicolumn{2}{@{}l}{\textbf{(b) Proof-Step Coverage}} \\
\addlinespace[1pt]
0 & Textbook proof not represented. The Lean proof does not contain the textbook's essential proof strategy or steps, even if it proves a related result. \\
1 & Minimal overlap with textbook proof. Only isolated pieces of the textbook argument appear; most essential proof moves are absent or replaced by unrelated reasoning. \\
2 & Some central proof steps are present, but multiple essential steps are missing, bypassed, or replaced by substantially different mathematical arguments. \\
3 & Mostly covered with significant gaps. The Lean proof includes the main proof strategy and several essential steps, but one or more important steps are absent, compressed beyond recoverability, or only indirectly represented. \\
4 & Nearly complete step coverage. All essential textbook steps are present, with only minor omissions or compressions that do not change the mathematical proof strategy. \\
5 & Every essential mathematical step of the textbook proof is represented in the Lean proof, allowing for routine formal elaboration and harmless decomposition into sublemmas. \\
\midrule
\multicolumn{2}{@{}l}{\textbf{(c) Proof-Structure Fidelity}} \\
\addlinespace[1pt]
0 & Incompatible proof structure. The formal proof follows a fundamentally different dependency structure from the textbook proof. \\
1 & Severe structural mismatch. The proof contains some textbook-like steps but combines or orders them in a way that changes the argument's mathematical logic. \\
2 & Partial structural fidelity. The broad strategy is recognizable, but important dependencies, branches, or contradiction structure are missing, reversed, or replaced. \\
3 & Mostly faithful structure with notable deviations. The proof follows the textbook's main dependency pattern, but one or more significant structural components are altered, hidden, or not justified in the same way. \\
4 & Faithful structure with minor formal differences. The proof preserves the textbook's dependency structure. Differences are limited to formal necessities, such as making implicit cases explicit or splitting bundled steps into lemmas. \\
5 & Fully faithful structure. The proof's mathematical dependencies match the textbook argument closely, including construction order, use of intermediate facts, branching, contradiction structure, and final conclusion. \\
\midrule
\multicolumn{2}{@{}l}{\textbf{(d) Cited-Dependency Fidelity}} \\
\addlinespace[1pt]
0 & Cited dependencies ignored or replaced wholesale. The formal proof does not use the textbook's cited mathematical dependencies or equivalent principles, and instead relies on unrelated results. \\
1 & Major dependency mismatch. Most important cited dependencies are absent, replaced by stronger later results, or used in ways that do not correspond to the textbook. \\
2 & Partial dependency fidelity. Some cited dependencies are represented, but several important ones are missing, mismatched, or replaced by non-equivalent principles. \\
3 & Mostly faithful dependencies with significant caveats. The main cited dependencies are present, but at least one important dependency is replaced by a stronger or less directly corresponding formal result. \\
4 & Faithful dependencies with minor formal substitutions. The proof uses formal analogues of the cited dependencies. Any substitutions are mathematically equivalent or are unavoidable due to the formal library's organization. \\
5 & Fully faithful dependencies. Each cited textbook dependency is represented by a direct formal analogue or a clearly equivalent formal principle used at the corresponding mathematical point in the proof. \\
\midrule
\multicolumn{2}{@{}l}{\textbf{(e) Assumption and Side-Condition Fidelity}} \\
\addlinespace[1pt]
0 & Unjustified assumptions central to the proof. The proof relies on assumptions or side conditions that are not in the textbook, not implied by the setup, and materially affect the result. \\
1 & Major assumption problems. Several important side conditions are unjustified, too strong, or inconsistent with the textbook configuration. \\
2 & Partial assumption fidelity. Some side conditions are justified, but important assumptions remain unsupported, stronger than what the textbook permits, or mathematically unclear. \\
3 & Mostly justified assumptions with notable concerns. Most added formal obligations follow from the textbook setup, but at least one nontrivial side condition is unjustified, uncertain, or requires a stronger reading of the textbook than warranted. \\
4 & Faithful assumptions with minor gaps. Added assumptions and side conditions are justified by the textbook setup or standard Euclidean background. Remaining issues are minor, local, or evidentially uncertain without threatening the proof's mathematical faithfulness. \\
5 & Fully faithful assumptions and side conditions. Every formal side condition used by the proof is justified by the textbook statement, the constructed configuration, prior established facts, or accepted Euclidean background principles. This score is compatible with automation when the automated obligations are mathematically justified. \\
\bottomrule
\end{tabular}
\caption{The five-dimension rubric shown to the LLM judge. Each dimension is scored independently on the same $0$--$5$ scale.}
\label{tab:llm-rubric}
\end{table*}

\section{Ablation Study Details}
\label{app:ablation-details}
This section details precisely how the ablation experiment was done. We compared the ablated arm, and full arm (\fillalgoname) using the same model Opus 4.8.

\paragraph{Isolation: two worktrees, one arm at a time.}
Each arm runs in its own git worktree on its own branch, because the arms require different repo contents:
The ablated branch removes the tools, hooks, and skills the full method (\fillalgoname) relies on,
so the two cannot share a working tree. We also \emph{never} run the two arms concurrently since models may update a memory file, so simultaneous runs could
cross-contaminate. Memory is wiped before every run and runs execute one arm at a time.
\paragraph{Prevent Cheating.}
The answers (fully proven propositions) exist in the git history, so agents could cheat. We could deny the raw usage of git in \texttt{settings.json}, but there are ways an agent can get around this, so instead we tell the agent what it is not allowed to do, and check suspicious behavior after the agent is done.

We specifically tell the agent that reading anything outside the repository or under the experiment folder, the
eval harness, or any use of \texttt{git}, auto-fails the attempt.

\paragraph{The prompts, and why the comparison is fair.}
Crucially, the ablated arm is \emph{not} handicapped by ignorance of what it must produce: its prompt
spells specifies how to output things, and what it is or is not allowed to do (for example that it cannot change the original formal claim).

The two task prompts are:
\begin{lstlisting}[basicstyle=\ttfamily\footnotesize,breaklines=true,numbers=none,xleftmargin=1em]
FULL (--my-method):
  Prove LeanEuclidF/<prop>/Main.lean end to end using the
  /faithful-prove skill...

ABLATED (--ablated):
  Prove LeanEuclidF/<prop>/Main.lean -- fill every ':= by sorry' so
  it builds with ZERO sorry. Do NOT change the theorem statement, the
  '(stepN : ...)' claim types, or the '-- @assumption (...)' lines.
  Anything Euclid cites must be cited in Lean too...

SHARED (both arms):
  Do NOT read anything outside this repository, or under
  reproducable_experiments/. Do NOT use git in any way. Either
  DISQUALIFIES the attempt -- AUTO-FAILED. No gaming the eval.
\end{lstlisting}

\paragraph{What the transcripts show.}
The ablated arm characteristically fails by attempting the whole proof at once, running a massive \texttt{lake build}, then rewriting files and re-firing another massive build. The full method
instead decomposes into small step lemmas built in fast scoped batches, making steady in-order progress.

\section{Compile Performance Details}
\label{app:compile-details}

The compile-time benchmark measures, per proposition, the wall-clock time to compile the faithful
decomposed proof (\texttt{Book1/PropN}) against the original proof
(\texttt{OldBook1/PropN}) done by \citet{murphy_autoformalizing_2024}; both prove the same \texttt{proposition\_N} and differ only in structure. The
procedure is: (1)~wipe \texttt{.lake/build} (the cached Mathlib packages are left untouched, so Mathlib is
never rebuilt); (2)~build \texttt{SystemE} once, untimed; (3)~compile each proposition cold, one at a
time, and record its wall-clock time, with a per-proposition $1$-hour cap. SMT is left uncapped
(\texttt{systemE.solverTime} raised to $100000$) so the $1$-hour wall is the only cutoff, and both experiments were ran on the same exclusive node (one after the other, not at the same time).
All runs used a single exclusive AMD~EPYC~9634 ($84$-core) node.

The ablation compile-time companion which evaluates the \emph{artifacts} the ablated, and \fillalgoname created (\cref{fig:ablation-compile}) follows the same procedure on an AMD~EPYC~9754 ($128$-core) node. The drivers, hardware metadata, and result CSVs are in the archive under \texttt{reproducable\_experiments/compile\_time/} (and \texttt{/ablation\_study/compile\_time/}).

\input{figures/fig-ablation-compile}

\section{Further Refutation Examples}
\label{app:further-refutation}
The main paper showed one example of a refutation. The released code contains further examples under \texttt{LeanEuclidF/accept\_refute/1.5/}. Importantly, it also contains a type B refutation example (only type A was shown in the main text).

Further, this folder contains example proofs that we \emph{accept} that differ from Euclid's own proof, showcasing that there are various ways to proove the same statement, to which our methodology can figure out which one to accept or refute formally.

\section{Limitation Details}
\label{app:limitations}

\paragraph{Skipped propositions.}
We complete $30$ of the $37$ propositions of Book~III. The remaining $7$ (Props.~24, 26--31) each require
extending the current implementation of System~E with a new primitive: an arc-length (circumference) measure for Props.~26--30, a
circle-superposition rule for Prop.~24, and a horn angle for Prop.~31. System~E is prior work
\citep{avigad2009formal}, implemented in Lean by \citet{murphy_autoformalizing_2024}, and adding such
primitives was decided to be out of scope.

\paragraph{Reliance on the oracle.}
\methodname is oracle-guided: a human (or automated) \oracle validates the \atomicity and
\atomicfaithfulness of the map stage. While LLM oracle is a valid option, the quality of proofs may degrade compared to using a human oracle, which is a limitation of the current system. However, our methodology still minimizes human-oracle effort so it is better than having the oracle complete the proof fully.
Nonetheless, reducing oracle use is a natural direction for future
work.

%% file: figures/fig-input-prop6.tex
\begin{figure*}[t]
\centering
\begin{lstlisting}[style=lean]
-- ============ formal(P_NL) : the Lean theorem  (Book1/Prop06/Main.lean) ============
theorem proposition_6 : ∀ (a b c : Point) (AB BC AC : Line),
  formTriangle a b c AB BC AC ∧ (∠ a:b:c = ∠ a:c:b)   -- hypothesis  α = α₁ ∧ ...
  → |(a─b)| = |(a─c)|                                  -- conclusion  β
--   free variables v₁..vₖ :  a b c,  AB BC AC

-- ============ P_NL : the natural-language proposition ============
-- "If a triangle has two angles equal to one another then the sides subtending
--  the equal angles will also be equal to one another."

-- ============ F_NL : the natural-language proof that gets segmented ============
-- "Let ABC be a triangle having the angle ABC equal to the angle ACB. I say that
--  side AB is also equal to side AC.  For if AB is unequal to AC then one of them
--  is greater.  Let AB be greater.  And let DB, equal to the lesser AC, have been
--  cut off from the greater AB [Prop. 1.3].  And let DC have been joined [Post. 1].
--  Therefore, since DB is equal to AC, and BC (is) common, the two sides DB, BC are
--  equal to the two sides AC, CB, respectively, and the angle DBC is equal to the
--  angle ACB.  Thus, the base DC is equal to the base AB, and the triangle DBC will
--  be equal to the triangle ACB [Prop. 1.4], the lesser to the greater.  The very
--  notion (is) absurd [C.N. 5].  Thus, AB is not unequal to AC.  Thus, (it is) equal."

-- ============ Δ : the fixed System E axiom set  (SystemE/Theory/**), e.g. ============
axiom two_points_determine_line : ∀ (a b : Point) (L M : Line),
  distinctPointsOnLine a b L ∧ (a.onLine M) ∧ (b.onLine M) → L = M
axiom segment_symmetric : ∀ (a b : Point), |(a─b)| = |(b─a)|
axiom angle_symm : ∀ (a b c : Point), (a ≠ b) ∧ (b ≠ c) → ((∠ a:b:c) = (∠ c:b:a))
axiom superposition : ∀ (a b c d g h : Point) (AB BC AC L : Line), …
-- …
\end{lstlisting}
\caption{The input tuple $\langle \propN, \formal{\propN}, \PropNProof, \systemEAxioms \rangle$
instantiated on Book~I, Prop.~6. Every component is reproduced verbatim from the released
artifact.}
\label{fig:input-prop6}
\end{figure*}

%% file: figures/fig-map-prop6.tex
\begin{figure*}[t]
\centering
\begin{lstlisting}[style=lean, breaklines=false,
  basicstyle=\ttfamily\fontsize{8}{8.6}\selectfont]
theorem proposition_6 : ∀ (a b c : Point) (AB BC AC : Line),
  formTriangle a b c AB BC AC ∧ (∠ a:b:c = ∠ a:c:b) → |(a─b)| = |(a─c)| := by
  euclid_intros
  have habsurd : ¬ (|(a─b)| ≠ |(a─c)|) := by
    intro hne
    -- s₁ = "For if AB is unequal to AC then one of them is greater."
    -- assert(s₁) = "one of them is greater"
    -- formal(assert(s₁)) = |(a─b)| > |(a─c)| ∨ |(a─c)| > |(a─b)|
    -- assumptions(s₁) = { "AB is unequal to AC" },  
    -- formal(a) = |(a─b)| ≠ |(a─c)|, where a is the one element in assumptions(s₁)
    -- @assumption ("AB is unequal to AC", |(a─b)| ≠ |(a─c)|)
    have step1_assumption1 : |(a─b)| ≠ |(a─c)| := by assumption  -- materialized assumption
    euclid_sentence "1.6.1" "For if AB is unequal to AC then one of them is greater."
      (step1 : |(a─b)| > |(a─c)| ∨ |(a─c)| > |(a─b)|) := by sorry
    wlog hgt : |(a─b)| > |(a─c)| generalizing b c AB BC AC with Hsym
    · have swapfig :                              -- another helper `have` in φ′
          (∠ a:c:b = ∠ a:b:c) ∧ (a ≠ c) ∧ (AC ≠ BC) ∧ (BC ≠ AB) ∧ (AB ≠ AC)
          ∧ (|(a─c)| ≠ |(a─b)|) ∧ (|(a─c)| > |(a─b)| ∨ |(a─b)| > |(a─c)|)
          ∧ (|(a─c)| > |(a─b)|) := by sorry
      obtain ⟨…⟩ := swapfig
      exact Hsym c b AC BC AB …
    · euclid_sentence "1.6.2" "Let AB be greater."
        (step2 : |(a─b)| > |(a─c)|) := by sorry
      ⋮
      euclid_sentence "1.6.9" "The very notion (is) absurd [C.N. 5]."
        (step9 : False) := by sorry
      exact step9
  euclid_sentence "1.6.10" "Thus, AB is not unequal to AC."
    (step10 : ¬ (|(a─b)| ≠ |(a─c)|)) := by sorry
  euclid_sentence "1.6.11" "Thus, (it is) equal."
    (step11 : |(a─b)| = |(a─c)|) := by sorry
  exact step11

-- ─────────── separately, a mid-proof "what-to-show" sentence (Book I Prop. 19) ───────────
-- euclid_wts announces a goal instead of asserting a fact, so:
--   assert(sᵢ) = "" (empty),  assumptions(sᵢ) = ∅,  formal(assert(sᵢ)) = True
  euclid_wts "1.19.2" "In fact, AC is not equal to AB."
\end{lstlisting}
\caption{Example of artifact at the end of \emph{map-stage}, $M=\langle \bar{\phi'}, \bar{s}, \assumptions{\cdot},
\assert{\cdot}, \formal{\cdot}\rangle$.
The \texttt{euclid\_sentence} tactic \emph{is} the mapping-object constructor: it emits the
ordered NL partition $\bar{s}=\{s_1,\dots,s_n\}$, and for each $s_i$ its type $(\texttt{step}i : \tau)$ is $\formal{\assert{s_i}}$
while the \texttt{-{}- @assumption} tags above it pinpoint $\assumptions{s_i}$ together with each
$\formal{a}$. Each \texttt{@assumption} is also materialized as a \texttt{have} (e.g.\
\texttt{step1\_assumption1}) by the assumption stage (\cref{sec:assumption-tagging}). A \texttt{euclid\_wts} is also shown at the end. The way to know if it is ``wts`` or not is by context of what is around it.}
\label{fig:map-prop6}
\end{figure*}

%% file: figures/fig-fill-prop6.tex
\begin{figure*}[t]
\centering
\begin{lstlisting}[style=lean, breaklines=false,
  basicstyle=\ttfamily\fontsize{8}{8.6}\selectfont]
-- BEFORE (map stage): step9 is left as `sorry` -- the missing κ₁,…,κₜ
  euclid_sentence "1.6.9" "The very notion (is) absurd [C.N. 5]."
    (step9 : False) := by sorry

-- AFTER (fill stage):  the sorry is replaced by a call to the backing lemma, which supplies κ₁,…,κₜ.
  euclid_sentence "1.6.9" "The very notion (is) absurd [C.N. 5]."
    (step9 : False) := by euclid_apply (helper_1_6_step9 a b c d AB BC AC …)

theorem helper_1_6_step9 (a b c d : Point) (AB BC AC : Line)
    (hbda : between b d a) (hbdac : |(b─d)| = |(a─c)|)
    (hdbc_acb : ∠ d:b:c = ∠ a:c:b)
    (harea : Triangle.area △d:b:c = Triangle.area △a:c:b) … : 
    False := by
  have hoff  : ¬ c.onLine AB := by euclid_finish                                -- one of the κᵢ
  have hdec  : Triangle.area △a:d:c + Triangle.area △c:d:b
                 = Triangle.area △a:c:b := by euclid_apply (sum_areas_if a b d c AB); assumption
  have hpos  : 0 < Triangle.area △a:d:c := by euclid_finish
  have hsymm : Triangle.area △c:d:b = Triangle.area △d:b:c := by euclid_finish
  linarith
\end{lstlisting}
\caption{The \fillstage on sentence \texttt{1.6.9} of Prop.~6. \textbf{Before}, the map stage leaves
$\texttt{step9}$ as \texttt{sorry}, the placeholder for the missing steps
$\kappa_1,\dots,\kappa_t$. \textbf{After}, the body after being filled becomes
\texttt{:= by euclid\_apply (helper\_1\_6\_step9 \dots)}}
\label{fig:fill-prop6}
\end{figure*}

%% file: figures/fig-birdseye-prop6.tex
\begin{figure*}[t]
\centering
\includegraphics[width=\textwidth]{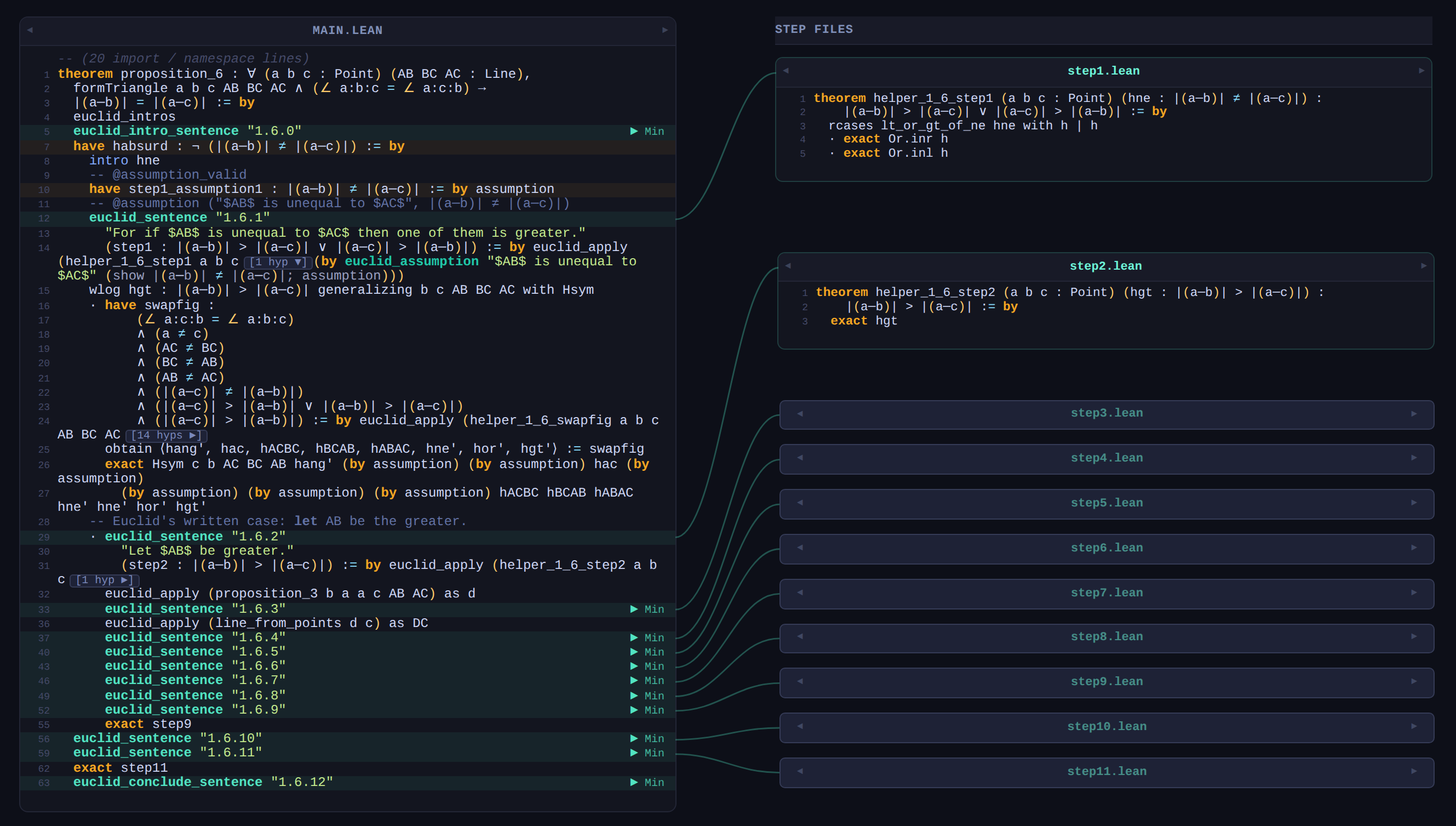}
\caption{Bird's-eye view of the completed Book~I Prop.~6 artifact. In the codebase, these are interactive html artifacts (\texttt{map.html}).
}
\label{fig:birdseye-prop6}
\end{figure*}

%% file: figures/fig-radar-glm.tex
\input{data/llm_judge/scores_glm5.tex}
\par\medskip
\begingroup\centering
\begin{minipage}{\linewidth}
\centering
\resizebox{0.85\linewidth}{!}{%
\tikz[baseline=-0.5ex]{%
  \draw[cRadarOursD,thick] (0,0) -- (0.5,0);
  \fill[cRadarOursD] (0.25,0) circle(1.4pt);
  \node[right,font=\scriptsize] at (0.55,0) {\methodname (Ours)};
  \draw[cRadarBaseD,thick,dashed] (2.3,0) -- (2.8,0);
  \fill[cRadarBaseD] (2.55,0) circle(1.4pt);
  \node[right,font=\scriptsize] at (2.85,0) {Baseline (LeanEuclid)};
  \draw[cRadarLLMD,thick,dashed] (5.5,0) -- (6.0,0);
  \fill[cRadarLLMD] (5.75,0) circle(1.4pt);
  \node[right,font=\scriptsize] at (6.05,0) {Baseline (Claude Code)};
}}\\[1pt]
\begin{minipage}[b]{0.26\linewidth}\centering
\radarpanel{\radarGlmNewOne}{\radarGlmOldOne}{3}{5}{cRadarBase}{cRadarBaseD}\\[-2pt]
{\scriptsize (a) Book~I}
\end{minipage}%
\hfill
\begin{minipage}[b]{0.26\linewidth}\centering
\radarpanel{\radarGlmNewTwo}{\radarGlmOldTwo}{0}{5}{cRadarLLM}{cRadarLLMD}\\[-2pt]
{\scriptsize (b) Book~II}
\end{minipage}%
\hfill
\begin{minipage}[b]{0.26\linewidth}\centering
\radarpanel{\radarGlmNewThree}{}{0}{5}{cRadarBase}{cRadarBaseD}\\[-2pt]
{\scriptsize (c) Book~III}
\end{minipage}
\captionof{figure}{%
  GLM-5 LLM-as-judge faithfulness by book (axes as in \cref{fig:llm-as-judge-radar}).
}
\label{fig:llm-as-judge-radar-glm}
\end{minipage}
\par\endgroup
\medskip

%% file: data/llm_judge/scores_glm5.tex
\def\radarGlmNewOne{4.88,4.90,4.85,4.94,4.85}
\def\radarGlmOldOne{4.15,4.29,4.17,4.58,4.23}
\def\radarGlmNewTwo{4.86,4.93,4.86,4.79,4.71}
\def\radarGlmOldTwo{3.50,1.57,1.43,1.36,4.57}
\def\radarGlmNewThree{4.83,4.73,4.70,4.83,4.77}
\def\radarGlmNOne{48}
\def\radarGlmNTwo{14}
\def\radarGlmNThree{30}

%% file: figures/fig-ablation-compile.tex

%% file: main.bbl
\begin{thebibliography}{32}
\providecommand{\natexlab}[1]{#1}

\bibitem[{Avigad, Dean, and Mumma(2009)}]{avigad2009formal}
Avigad, J.; Dean, E.; and Mumma, J. 2009.
\newblock A Formal System for {Euclid's} \emph{Elements}.
\newblock \emph{The Review of Symbolic Logic}, 2(4): 700–768.

\bibitem[{Azerbayev et~al.(2023)Azerbayev, Piotrowski, Schoelkopf, Ayers,
  Radev, and Avigad}]{azerbayev2023proofnet}
Azerbayev, Z.; Piotrowski, B.; Schoelkopf, H.; Ayers, E.~W.; Radev, D.; and
  Avigad, J. 2023.
\newblock {ProofNet}: Autoformalizing and Formally Proving Undergraduate-Level
  Mathematics.
\newblock \emph{arXiv preprint}.

\bibitem[{Beeson, Narboux, and Wiedijk(2018)}]{beeson2019proof}
Beeson, M.; Narboux, J.; and Wiedijk, F. 2018.
\newblock Proof-checking Euclid.
\newblock arXiv:1710.00787.

\bibitem[{Cabral et~al.(2025)Cabral, Do, Yu, Tai, Feng, and Shen}]{proofflow}
Cabral, R.; Do, T.~M.; Yu, X.; Tai, W.~M.; Feng, Z.; and Shen, X. 2025.
\newblock ProofFlow: A Dependency Graph Approach to Faithful Proof
  Autoformalization.
\newblock arXiv:2510.15981.

\bibitem[{Carneiro(2019)}]{carneiro2019typetheory}
Carneiro, M. 2019.
\newblock \emph{The Type Theory of {Lean}}.
\newblock Master's thesis, Carnegie Mellon University.
\newblock \url{https://github.com/digama0/lean-type-theory}.

\bibitem[{de~Moura and Ullrich(2021)}]{demoura2021lean}
de~Moura, L.; and Ullrich, S. 2021.
\newblock {The Lean 4 Theorem Prover and Programming Language}.
\newblock In Platzer, A.; and Sutcliffe, G., eds., \emph{Automated Deduction
  (CADE 28)}, volume 12699 of \emph{Lecture Notes in Computer Science},
  625--635. Berlin, Heidelberg: Springer.

\bibitem[{Feng et~al.(2026)Feng, Pu, An, Bastani, Zhang, Huang, Si, and
  Li}]{feng2026lcsbench}
Feng, Y.; Pu, F.; An, O.; Bastani, O.; Zhang, L.; Huang, J.; Si, X.; and Li, Z.
  2026.
\newblock Theory-Scale Auto-Formalization of Logics for Computer Science.
\newblock arXiv:2606.26525.

\bibitem[{Fitzpatrick(2007)}]{fitzpatrick_rfitzpelements_2026}
Fitzpatrick, R. 2007.
\newblock Euclid's {Elements} of {Geometry} (Greek/English edition).
\newblock \url{https://github.com/rfitzp/Elements}.
\newblock Greek text after J.~L.~Heiberg (1883--1885); commit 5a6699e, accessed
  2026-07-28.

\bibitem[{Gonthier(2008)}]{gonthier2008four}
Gonthier, G. 2008.
\newblock {Formal Proof---The Four-Color Theorem}.
\newblock \emph{Notices of the American Mathematical Society}, 55(11):
  1382--1393.

\bibitem[{Hales et~al.(2017)Hales, Adams, Bauer, Dang, Harrison, Hoang,
  Kaliszyk, Magron, McLaughlin, Nguyen et~al.}]{hales2017formal}
Hales, T.; Adams, M.; Bauer, G.; Dang, T.~D.; Harrison, J.; Hoang, L.~T.;
  Kaliszyk, C.; Magron, V.; McLaughlin, S.; Nguyen, T.~T.; et~al. 2017.
\newblock {A Formal Proof of the Kepler Conjecture}.
\newblock \emph{Forum of Mathematics, Pi}, 5: e2.

\bibitem[{Hu, Zhu, and Welleck(2025)}]{hu2024minictx}
Hu, J.; Zhu, T.; and Welleck, S. 2025.
\newblock miniCTX: Neural Theorem Proving with (Long-)Contexts.
\newblock arXiv:2408.03350.

\bibitem[{Huang et~al.(2021)Huang, Li, Chen, Samel, Naik, Song, and
  Si}]{huang2021scallop}
Huang, J.; Li, Z.; Chen, B.; Samel, K.; Naik, M.; Song, L.; and Si, X. 2021.
\newblock {S}callop: From Probabilistic Deductive Databases to Scalable
  Differentiable Reasoning.
\newblock In \emph{Advances in Neural Information Processing Systems
  (NeurIPS)}.

\bibitem[{Jiang et~al.(2023)Jiang, Welleck, Zhou, Li, Liu, Jamnik, Lacroix, Wu,
  and Lample}]{jiang2023draft}
Jiang, A.~Q.; Welleck, S.; Zhou, J.~P.; Li, W.; Liu, J.; Jamnik, M.; Lacroix,
  T.; Wu, Y.; and Lample, G. 2023.
\newblock Draft, Sketch, and Prove: Guiding Formal Theorem Provers with
  Informal Proofs.
\newblock arXiv:2210.12283.

\bibitem[{Lample et~al.(2022)Lample, Lachaux, Lavril, Martinet, Hayat, Ebner,
  Rodriguez, and Lacroix}]{lample2022hypertree}
Lample, G.; Lachaux, M.-A.; Lavril, T.; Martinet, X.; Hayat, A.; Ebner, G.;
  Rodriguez, A.; and Lacroix, T. 2022.
\newblock {HyperTree} Proof Search for Neural Theorem Proving.
\newblock \emph{arXiv preprint}.

\bibitem[{Li, Huang, and Naik(2023)}]{li2023scallop}
Li, Z.; Huang, J.; and Naik, M. 2023.
\newblock {S}callop: A Language for Neurosymbolic Programming.
\newblock \emph{Proceedings of the ACM on Programming Languages (PLDI)}, 7.

\bibitem[{Li et~al.(2024)Li, Sun, Murphy, Su, Li, Zhang, Yang, and
  Si}]{li2024survey}
Li, Z.; Sun, J.; Murphy, L.; Su, Q.; Li, Z.; Zhang, X.; Yang, K.; and Si, X.
  2024.
\newblock A Survey on Deep Learning for Theorem Proving.
\newblock \emph{arXiv preprint}.

\bibitem[{Li et~al.(2026)Li, Yang, He, Zhao, Zhao, Tang, Yang, Gupta, Su, and
  Jin}]{goedelcodeprover}
Li, Z.; Yang, Z.; He, D.; Zhao, H.; Zhao, A.; Tang, S.; Yang, K.; Gupta, A.;
  Su, Z.; and Jin, C. 2026.
\newblock Goedel-Code-Prover: Hierarchical Proof Search for Open
  State-of-the-Art Code Verification.
\newblock arXiv:2603.19329.

\bibitem[{Lu et~al.(2024)Lu, Wan, Huang, Xiong, Liu, and
  Guo}]{lu2024formalalign}
Lu, J.; Wan, Y.; Huang, Y.; Xiong, J.; Liu, Z.; and Guo, Z. 2024.
\newblock {FormalAlign}: Automated Alignment Evaluation for Autoformalization.
\newblock \emph{arXiv preprint}.

\bibitem[{Manhaeve et~al.(2018)Manhaeve, Dumancic, Kimmig, Demeester, and
  De~Raedt}]{manhaeve2018deepproblog}
Manhaeve, R.; Dumancic, S.; Kimmig, A.; Demeester, T.; and De~Raedt, L. 2018.
\newblock {D}eep{P}rob{L}og: Neural Probabilistic Logic Programming.
\newblock In \emph{Advances in Neural Information Processing Systems
  (NeurIPS)}.

\bibitem[{mathlib Community(2020)}]{mathlib2020}
mathlib Community, T. 2020.
\newblock The lean mathematical library.
\newblock In \emph{Proceedings of the 9th ACM SIGPLAN International Conference
  on Certified Programs and Proofs}, CPP ’20, 367–381. ACM.

\bibitem[{Min et~al.(2026)Min, Gao, Sy, Li, Si, and Bastani}]{min2026divide}
Min, M.~J.; Gao, Y.; Sy, W.; Li, Z.; Si, X.; and Bastani, O. 2026.
\newblock {Divide and Abstract: Autoformalization via Decomposition and
  Abstraction Learning}.
\newblock In \emph{International Conference on Learning Representations
  (ICLR)}.

\bibitem[{Murphy et~al.(2024)Murphy, Yang, Sun, Li, Anandkumar, and
  Si}]{murphy_autoformalizing_2024}
Murphy, L.; Yang, K.; Sun, J.; Li, Z.; Anandkumar, A.; and Si, X. 2024.
\newblock Autoformalizing {Euclidean} {Geometry}.
\newblock ArXiv:2405.17216 [cs.LG].

\bibitem[{Poiroux et~al.(2025)Poiroux, Weiss, Kun{\v{c}}ak, and
  Bosselut}]{poiroux2025reliable}
Poiroux, A.; Weiss, G.; Kun{\v{c}}ak, V.; and Bosselut, A. 2025.
\newblock Reliable Evaluation and Benchmarks for Statement Autoformalization.
\newblock In \emph{Conference on Empirical Methods in Natural Language
  Processing (EMNLP)}.

\bibitem[{Tao(2026)}]{tao2026ai}
Tao, T. 2026.
\newblock {Mathematics in the Age of AI}.
\newblock Public lecture, International Congress of Mathematicians.

\bibitem[{Thurston(1994)}]{thurston1994proof}
Thurston, W.~P. 1994.
\newblock {On Proof and Progress in Mathematics}.
\newblock \emph{Bulletin of the American Mathematical Society}, 30(2):
  161--177.

\bibitem[{Turpin et~al.(2023)Turpin, Michael, Perez, and
  Bowman}]{turpin2023language}
Turpin, M.; Michael, J.; Perez, E.; and Bowman, S.~R. 2023.
\newblock Language Models Don't Always Say What They Think: Unfaithful
  Explanations in Chain-of-Thought Prompting.
\newblock arXiv:2305.04388.

\bibitem[{Wang et~al.(2025)Wang, Li, Song, Xu, Tang, Zhuge, Pan, Song, Li,
  Singh, Tran, Li, Ma, Zheng, Qian, Shao, Muennighoff, Zhang, Hui, Lin,
  Brennan, Peng, Ji, and Neubig}]{wang2025openhands}
Wang, X.; Li, B.; Song, Y.; Xu, F.~F.; Tang, X.; Zhuge, M.; Pan, J.; Song, Y.;
  Li, B.; Singh, J.; Tran, H.~H.; Li, F.; Ma, R.; Zheng, M.; Qian, B.; Shao,
  Y.; Muennighoff, N.; Zhang, Y.; Hui, B.; Lin, J.; Brennan, R.; Peng, H.; Ji,
  H.; and Neubig, G. 2025.
\newblock {OpenHands}: An Open Platform for {AI} Software Developers as
  Generalist Agents.
\newblock In \emph{International Conference on Learning Representations
  (ICLR)}.

\bibitem[{Wu et~al.(2022)Wu, Jiang, Li, Rabe, Staats, Jamnik, and
  Szegedy}]{wu2022autoformalization}
Wu, Y.; Jiang, A.~Q.; Li, W.; Rabe, M.~N.; Staats, C.; Jamnik, M.; and Szegedy,
  C. 2022.
\newblock Autoformalization with Large Language Models.
\newblock In \emph{Advances in Neural Information Processing Systems
  (NeurIPS)}.

\bibitem[{Yang et~al.(2023)Yang, Swope, Gu, Chalamala, Song, Yu, Godil,
  Prenger, and Anandkumar}]{yang2023leandojo}
Yang, K.; Swope, A.~M.; Gu, A.; Chalamala, R.; Song, P.; Yu, S.; Godil, S.;
  Prenger, R.; and Anandkumar, A. 2023.
\newblock {LeanDojo}: Theorem Proving with Retrieval-Augmented Language Models.
\newblock In \emph{Advances in Neural Information Processing Systems (NeurIPS)
  Datasets and Benchmarks Track}.

\bibitem[{Yu et~al.(2025)Yu, Zhong, Feng, Zhai, Yousefzadeh, Ng, Liu, Shou,
  Xiong, Zhou, Ong, Sugiarto, Zhang, Tai, Cao, Lu, Sun, Xu, Xin, and
  Li}]{yu2025mathesis}
Yu, X.; Zhong, J.; Feng, Z.; Zhai, P.; Yousefzadeh, R.; Ng, W.~C.; Liu, H.;
  Shou, Z.; Xiong, J.; Zhou, Y.; Ong, C.~B.; Sugiarto, A.~J.; Zhang, Y.; Tai,
  W.~M.; Cao, H.; Lu, D.; Sun, J.; Xu, Q.; Xin, S.; and Li, Z. 2025.
\newblock Mathesis: Towards Formal Theorem Proving from Natural Languages.
\newblock arXiv:2506.07047.

\bibitem[{Zheng et~al.(2024)Zheng, Wang, Xie, Liu, Sun, Xin, Shen, Li, and
  Li}]{zheng2024lyra}
Zheng, C.; Wang, H.; Xie, E.; Liu, Z.; Sun, J.; Xin, H.; Shen, J.; Li, Z.; and
  Li, Y. 2024.
\newblock Lyra: Orchestrating Dual Correction in Automated Theorem Proving.
\newblock arXiv:2309.15806.

\bibitem[{Zheng, Han, and Polu(2022)}]{zheng2022minif2f}
Zheng, K.; Han, J.~M.; and Polu, S. 2022.
\newblock {MiniF2F}: a cross-system benchmark for formal {O}lympiad-level
  mathematics.
\newblock In \emph{International Conference on Learning Representations
  (ICLR)}.

\end{thebibliography}
